\documentclass[11pt]{article}

\usepackage[final]{acl}

\usepackage{times}
\usepackage{latexsym}

\usepackage[T1]{fontenc}

\usepackage[utf8]{inputenc}

\usepackage{microtype}

\usepackage{inconsolata}

\usepackage{graphicx}
\usepackage{amssymb}

\usepackage{enumitem}
\usepackage{multirow}
\usepackage{makecell}
\usepackage[table]{xcolor}
\usepackage{booktabs}
\usepackage{tabularx}
\usepackage{amsmath}

\title{MODE: Modality-Decomposed Expert-Level Mixed-Precision \\
Quantization for MoE Multimodal LLMs}

\author{
 \textbf{Yuanteng Chen\textsuperscript{1,2,3}} \thanks{$\ $  Equal contribution.},
 \textbf{Nanxin Zeng\textsuperscript{2}} \footnotemark[1],
 \textbf{Peisong Wang\textsuperscript{1,2}} \thanks{$\ $  Corresponding author.},
 \textbf{Zhilei Liu\textsuperscript{1,2}},
\\
 \textbf{Yuantian Shao\textsuperscript{1,5}},
 \textbf{Shiqiang Lang\textsuperscript{3}},
 \textbf{Tao Liu\textsuperscript{3}},
 \textbf{Chuangyi Li\textsuperscript{1,2}},
\\
 \textbf{Qinghao Hu\textsuperscript{1,2}},
 \textbf{Gang Li\textsuperscript{1,2}},
 \textbf{Jing Liu\textsuperscript{1,2,3}},
 \textbf{Jian Cheng\textsuperscript{1,2,3}} \footnotemark[2] ,
\\
 \textsuperscript{1}Institute of Automation, Chinese Academy of Sciences,
 \\
 \textsuperscript{2}School of Artificial Intelligence, University of Chinese Academy of Sciences,
 \\
 \textsuperscript{3}Zhongguancun Acedemy,
 \textsuperscript{4}School of Computer Science, University of Sydney,
 \\
 \textsuperscript{5}Nanjing University of Science and Technology
\\
 \small{
   \textbf{Correspondence:} \texttt{\{peisong.wang, jcheng\}@nlpr.ia.ac.cn}
 }
}

\begin{document}
\maketitle
\begin{abstract}
Mixture-of-Experts Multimodal Large Language Models (MoE-MLLMs) offer remarkable performance but incur prohibitive GPU memory costs, making compression essential. Among PTQ methods, expert-level mixed-precision quantization has proven effective for MoE-LLMs, yet suffers notable degradation on MoE-MLLMs due to two overlooked biases in expert importance estimation. (1) At the \textbf{cross-modal} level, the numerical dominance of vision tokens causes expert selection frequency to be dominated by vision tokens, masking experts that are critical to the text modality; (2) at the \textbf{intra-vision} level, the large proportion of redundant vision tokens further skew frequency statistics, obscuring experts critical for informative visual content. To bridge gaps, we propose \textbf{MODE}, a \textbf{mo}dality-\textbf{d}ecomposed \textbf{e}xpert-level mixed-precision quantization framework for MoE-MLLMs that decomposes expert selection frequency by modality, filters redundant vision tokens to obtain denoised visual frequency, and further evaluates quantization sensitivity per modality as a complementary signal to frequency-based estimation. These signals are integrated into an Integer Linear Programming formulation to assign per-expert bit-widths under a given budget. 
Extensive experiments show that MODE is particularly well-suited for MoE-MLLMs, limiting average performance loss to within 2.9\% at W3A16, with larger gains at the extreme 2-bit setting. The code is available at \href{https://github.com/MingZwhy/MODE}{Github}.
\end{abstract}

\section{Introduction}

\begin{figure}[t]
  \centering
  \includegraphics[width=\columnwidth]{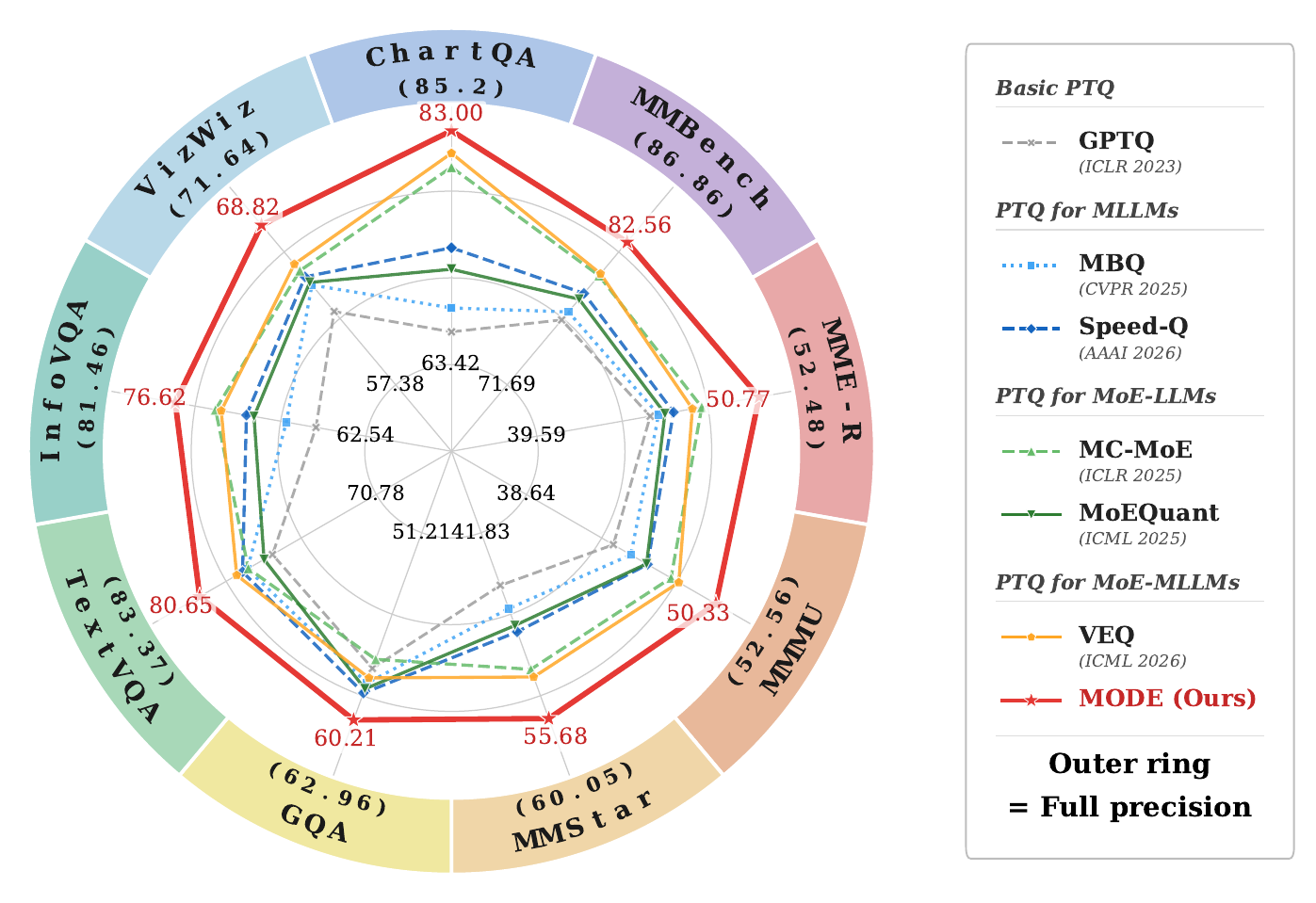}
  \caption{Performance comparison on Qwen3-VL-30B-A3B-Instruct under 3-bit weight quantization (W3A16).}
  \label{fig:radar_w3}
  \vskip -0.15in
\end{figure}

Multimodal Large Language Models (MLLMs) have achieved remarkable success across a wide range of vision-language tasks~\citep{pmlr-v139-radford21a,vteam2026glm45vglm41vthinkingversatilemultimodal}. By aligning visual perception with linguistic reasoning, they empower intelligent systems to perceive, understand, and interact with the visual world in a unified framework.

With the growing scale of MLLMs, the Mixture-of-Experts (MoE) architecture~\citep{artetxe-etal-2022-efficient} has become a popular solution to manage computational cost, activating only a sparse subset of experts per token to enable efficient parameter scaling while keeping training and inference FLOPs low~\citep{fedus2022switchtransformersscalingtrillion}. However, since all expert parameters must reside on GPUs before inference regardless of activation sparsity, deploying large-scale MoE-MLLMs~\citep{lin2024moellavamixtureexpertslarge} like Qwen3-VL-30B-A3B-Instruct~\citep{qwen3technicalreport} still demands prohibitive memory, which severely limits their practical use in resource-constrained settings.

Post-Training Quantization (PTQ) offers a practical way to reduce memory footprint without retraining, yet existing PTQ methods target either dense MLLMs or MoE-LLMs, and neither transfers well to MoE-MLLMs. Specifically, PTQ methods designed for dense MLLMs~\citep{li2024mbq,guo2025speedq} recognize the modality gap between text and vision tokens, but treat the model as a monolithic whole and overlook the sparse activation and unequal expert contribution intrinsic to MoE architectures. Conversely, PTQ methods tailored for MoE~\citep{zheng2026dynamoruntimeswitchablequantization,hu2025moequantenhancingquantizationmixtureofexperts} exploit these structural properties, but 
ignore the modality differences inherent to MLLMs.

Among PTQ methods for MoE, expert-level mixed-precision quantization~\citep{huang2025mixturecompressormixtureofexpertsllms} has emerged as a prevailing paradigm, which measures expert importance through activation frequency and assigns higher bit-widths to more critical experts while aggressively compressing less important ones. However, when applied to MoE-MLLMs, these methods aggregate activation statistics across all tokens regardless of modality, rendering expert importance estimates unreliable and leading to notable performance degradation.

In our study, we identify two distinct levels of bias that arise when expert-level mixed-precision quantization is applied to MoE-MLLMs. At the cross-modal level, vision tokens vastly outnumber text tokens in typical multimodal inputs, so global frequency statistics are dominated by vision-side routing patterns and systematically undervalue experts critical for textual reasoning. At the intra-vision level, pervasive redundancy among vision tokens further distorts frequency-based importance, as many tokens carry near-duplicate information yet each still casts a router vote, inflating the apparent importance of certain experts while obscuring the contributions of others.

Motivated by these observations, we propose \textbf{MODE}, a modality-aware quantization framework tailored to MoE-MLLMs that jointly leverages the heterogeneous expert structure of MoE for differentiated precision allocation and accounts for the modality heterogeneity and intra-vision redundancy inherent in multimodal inputs. To resolve the cross-modal bias, MODE separately collects and normalizes expert selection frequencies for text and vision tokens, so that text-critical experts are no longer overshadowed. To address the intra-vision distortion, MODE identifies key vision tokens at each layer and computes vision-side frequencies solely from them, filtering out noise from redundant visual content. Beyond frequency alone, MODE further evaluates each expert's quantization sensitivity separately under text and key vision tokens, yielding a modality-decomposed importance metric. The resulting scores are integrated into an Integer Linear Programming (ILP) formulation that 
produces an optimal per-expert precision allocation under a given bit budget.

We evaluate MODE on three MoE-MLLM families across ten multimodal benchmarks under various low-bit weight-only quantization settings. As illustrated in Figure~\ref{fig:radar_w3}, MODE consistently surpasses all compared PTQ methods designed for dense MLLMs, MoE-LLMs, and MoE-MLLMs, respectively. In particular, at the W3A16 setting MODE incurs only 2.84\% and 2.08\% average accuracy degradation on Qwen3-VL-30B-A3B-Instruct and Kimi-VL-A3B-Instruct, respectively, making practical deployment of 30B-scale MoE-MLLMs on a single consumer-grade GPU feasible.

\begin{figure*}[htb]
    \centering
    \includegraphics[width=\textwidth]{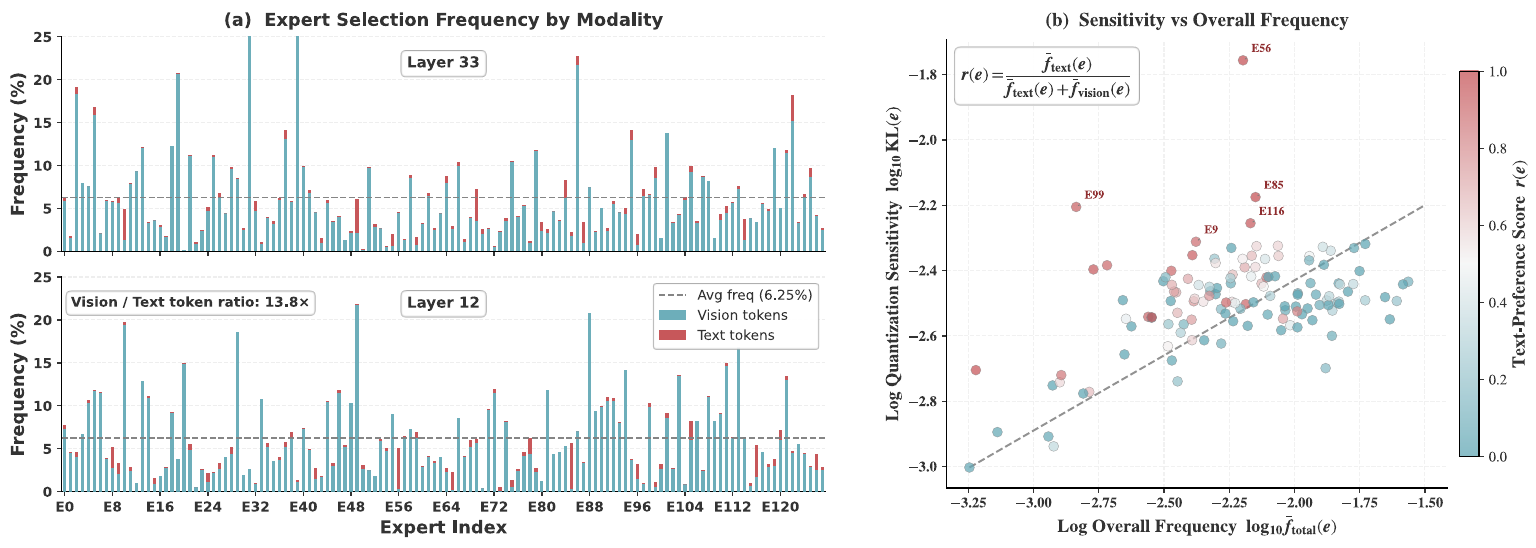}
    \vskip -0.1in
    \caption{Cross-modal expert frequency bias in Qwen3-VL-30B-A3B-Instruct. 
    (a) Per-expert selection frequency at Layers 12 and 33, decomposed into vision-token (blue) and text-token (red) contributions. (b) Quantization sensitivity (KL divergence between first output-token logits before and after 2-bit RTN quantization, log-scaled) versus overall selection frequency for all Layer-12 experts, colored by the text-preference score $r(e)$.}
    \label{fig:cross-modal-bias}
    \vskip -0.1in
\end{figure*}


\section{Related Work}
\noindent\textbf{MLLMs Quantization.}
PTQ methods designed for MLLMs primarily focus on mitigating the distribution heterogeneity between vision and text modalities. SPEED-Q~\citep{guo2025speedq} adopts a staged quantization strategy that first quantizes the vision encoder and re-calibrates the projection layer for modality alignment.
MBQ~\citep{li2024mbq} recognizes that vision and language tokens exhibit different quantization sensitivities and introduces gradient-based sensitivity measures into calibration to balance reconstruction quality across modalities.

\noindent\textbf{MoE Quantization.}
The sparse activation nature of MoE architectures
brings new challenges and opportunities to PTQ. MoEQuant~\citep{hu2025moequantenhancingquantizationmixtureofexperts} addresses inter- and intra-expert activation imbalance through an expert-balanced calibration strategy. A more prevalent line of work exploits the heterogeneous expert contributions through mixed-precision quantization. MC-MoE~\citep{huang2025mixturecompressormixtureofexpertsllms} 
leverages expert activation frequency as an importance indicator, assigning higher bit-widths to more critical experts. 
MoQa~\citep{zheng2026dynamoruntimeswitchablequantization} builds upon this paradigm by introducing a channel-level dynamic adjustment mechanism. 
More recently, VEQ~\citep{qin2026veqmodalityadaptivequantizationmoe} takes a first step toward MoE-MLLMs by incorporating a modality-affinity-aware Hessian objective based on token–expert affinity to improve accuracy. 

\section{Motivation}
\label{sec:motivation}
Expert selection frequency, defined as the fraction of tokens routed to a given expert on a calibration set, is a natural and widely adopted importance proxy for mixed-precision quantization, as more frequently activated experts are considered more important in MoE. However, when transferred to MoE-MLLMs, we identify two distinct levels of bias that cause overall selection frequency to significantly deviate from true expert importance. We elaborate on each below, using Qwen3-VL-30B-A3B-Instruct as the representative model and 512 randomly sampled image-text pairs from ShareGPT4V~\citep{chen2023sharegpt4v} as calibration set.

\subsection{Cross-Modal Expert Frequency Bias.}
\label{sec:cross-modal-bias}
In typical multimodal inputs, a single image is encoded into hundreds or even thousands of visual tokens while the accompanying text prompt remains relatively short, so vision tokens outnumber text tokens by a large margin. This imbalance is an intrinsic property of multimodal inference, and consequently the overall expert selection frequency is dominated by vision-side routing patterns, causing experts that are heavily selected by text tokens but receive little vision traffic to be systematically undervalued under a global frequency ranking. To illustrate this, Figure~\ref{fig:cross-modal-bias}(a) decomposes the per-expert selection frequency at two representative MoE layers into vision-token (blue) and text-token (red) contributions. Clear variation across experts confirms that experts do differ substantially in importance, supporting the general premise of frequency-based estimation; however, the text-token contribution is nearly invisible at this scale, and experts like E56 and E85 in Layer 12, despite being among the most heavily selected by text tokens, fall well below the layer average in total frequency and thus become inconspicuous.


Beyond frequency statistics, we further reveal this bias from the perspective of actual quantization loss. Figure~\ref{fig:cross-modal-bias}(b) plots each Layer-12 expert's quantization sensitivity, measured as the KL divergence between the first output-token logits before and after 2-bit RTN quantization (log-scaled, averaged over all calibration samples), against its log overall frequency, with redder color indicating experts more frequently selected by text tokens but less by vision tokens. At the global level, a broadly positive trend between sensitivity and frequency is visible, supporting frequency as a reasonable importance proxy in general. At the individual expert level, however, several strongly text-preferred experts emerge as pronounced outliers: E56 and E99 exhibit disproportionately high quantization sensitivity despite having only low-to-moderate overall frequencies, indicating that they should be protected as text-critical experts rather than being dismissed by their modest overall activation.

These observations motivate a straightforward design principle: text-token and vision-token frequencies should be normalized independently before being combined into a unified importance score, ensuring that experts critical to each modality receive appropriate protection.



\begin{figure}[hbt]
  \centering
  \includegraphics[width=\columnwidth]{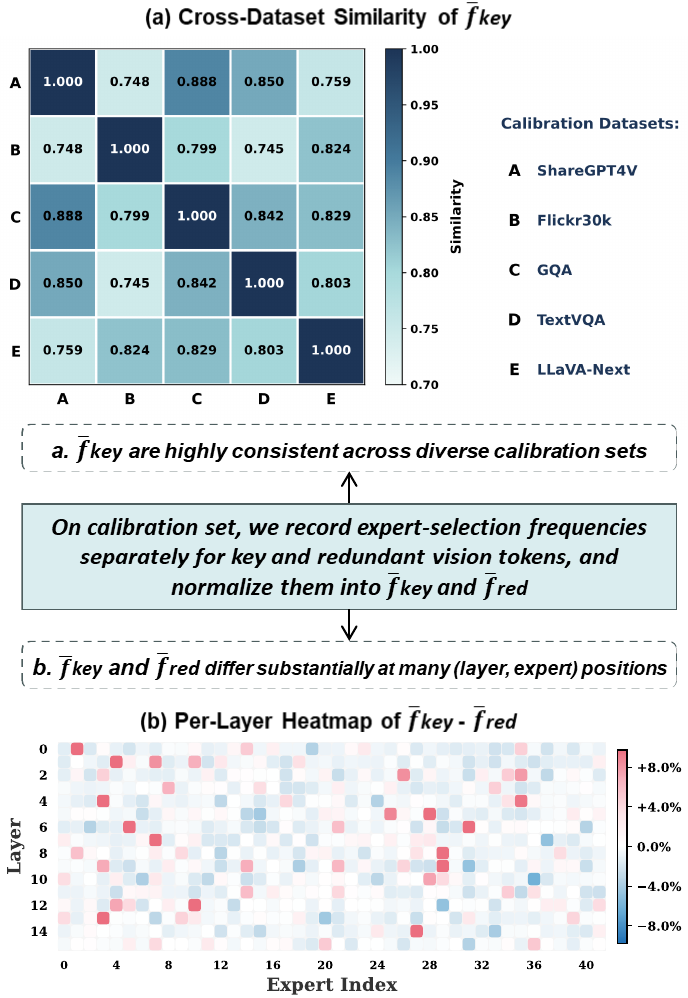}
  \vskip -0.1in
  \caption{Intra-vision expert frequency bias. 
  \textbf{(a)} $\bar{f}_{\mathrm{key}}$ is highly consistent across five calibration datasets. 
  \textbf{(b)} Within a single dataset, $\bar{f}_{\mathrm{key}}$ and $\bar{f}_{\mathrm{red}}$ exhibits pronounced deviations across (layer, expert) positions, revealing a systematic key–redundant routing bias.}
  \label{fig:intra-vision-bias}
  \vskip -0.1in
\end{figure}

\subsection{Intra-Vision Expert Frequency Bias}
\label{sec:intra-vision-bias}



Recent work on MLLM token compression~\citep{yang2026visionziplongerbetternecessary} consistently shows that only a small subset of vision tokens carry the core visual semantics, while the vast majority are redundant and can be pruned with negligible performance loss. In the context of MoE-MLLMs, however, a more pertinent question is whether these key vision tokens correspond to a distinct and stable group of experts---those that are truly important for visual modality. Our analysis gives an affirmative answer and further uncovers a second layer of frequency bias within vision modality itself: key and redundant vision tokens systematically activate different experts, while the preference of key tokens remains stable across diverse data, as detailed below.


To distinguish key vision tokens from redundant ones, we follow the widely adopted attention-based criterion of SparseVLM~\citep{zhang2024sparsevlm}: at each layer, we rank vision tokens by the total attention they receive from all text tokens, which reliably indicates their relevance to the current query, and treat the top 20\% as key tokens and the remainder as redundant. We apply this selection 
at every layer, so that the key-token set adapts to how the model attends to visual content at different depths. A detailed description is provided in Appendix~\ref{app:key-vision-token}.

With key and redundant vision tokens identified at every layer, we collect their layer-wise normalized expert-selection frequencies, $\bar{f}_{\mathrm{key}}$ and $\bar{f}_{\mathrm{red}}$, on calibration datasets and analyze from the following two complementary angles.

We first examine the cross-dataset behavior of $\bar{f}_{\mathrm{key}}$ via its pairwise cosine similarity across the five datasets (details in Appendix~\ref{app:freq-details}). As shown in Figure~\ref{fig:intra-vision-bias}(a), the scores are consistently high (all above 0.74), indicating that key vision tokens converge to a stable set of preferred experts regardless of data composition. This stability links key vision tokens to a group of vision-critical experts intrinsic to the model, rather than to dataset-specific artifacts.

We then zoom into a single calibration set (ShareGPT4V) to ask whether redundant tokens share the same preferences. Figure~\ref{fig:intra-vision-bias}(b) visualizes $\bar{f}_{\mathrm{key}} - \bar{f}_{\mathrm{red}}$ over a representative subset of (layer, expert) positions (the full 48$\times$128 grid is omitted for space), where a clear divergence emerges: certain experts are substantially more activated by key tokens, while others are disproportionately used by redundant ones. Since redundant tokens dominate the vision-token population, treating all vision tokens indiscriminately biases the estimated expert importance toward redundant-preferred experts, leaving those most relevant to meaningful visual content under-protected.

Combining the two findings, expert importance for vision modality should be assessed through lens of key vision tokens rather than all vision tokens.

\begin{figure}[b]
    \centering
    \includegraphics[width=\linewidth]{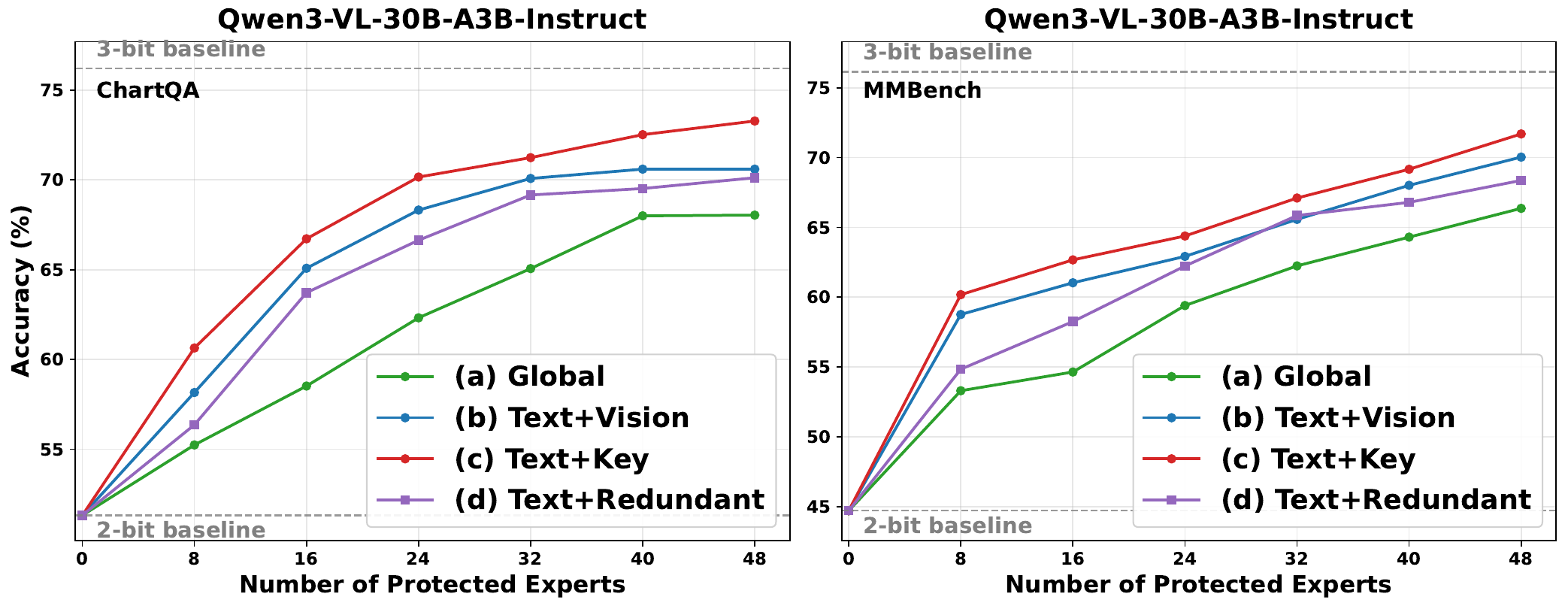}
    \caption{Concept verification. Experts are progressively promoted from 2-bit to 3-bit under four ranking strategies (a)--(d); results on ChartQA and MMBench validate both dimensions of frequency bias.}
    \label{fig:concept-verify}
    \vskip -0.1in
\end{figure}

\begin{figure*}[htb]
    \centering
    \includegraphics[width=\textwidth]{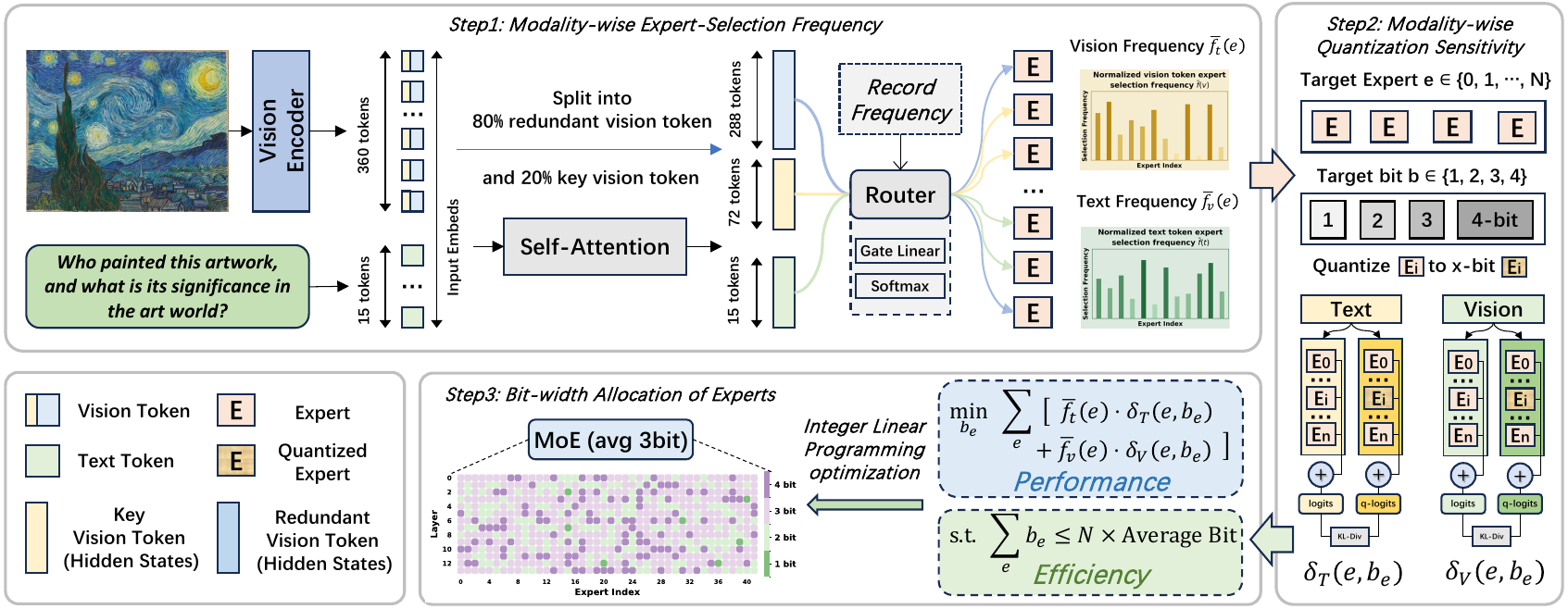}
    \caption{Pipeline of our quantization method for MoE-MLLMs. Modality-wise frequency and sensitivity are profiled per expert, then combined in an ILP that allocates expert-level bit-widths under a target budget.}
    \label{fig:pipeline}
    \vskip -0.1in
\end{figure*}



\subsection{Concept Verification}
\label{sec:concept_verification}
To jointly validate the two dimensions of bias, we conduct a controlled mixed-precision experiment on Qwen3-VL-30B-A3B-Instruct. We fix all attention layers at 4-bit precision and quantize every MoE expert to 2 bits via RTN as the starting point. We then incrementally promote experts to 3 bits in steps of 8 per layer (out of 128), where the order of promotion is determined by a \emph{simplified expert importance score} $s$ built from per-layer expert selection frequencies $\bar{f}_{x}$ (normalized over experts) on different token groups $x$. We compare four strategies for computing $s$, differing only in which token-group frequencies are used:

\vspace{2pt}
\noindent\textbf{(a) Global:}\quad $s = \bar{f}_{\mathrm{total}}$, the selection frequency over all calibration tokens.

\noindent\textbf{(b) Text+Vision:}\quad $s = \tfrac{1}{2} \bar{f}_{\mathrm{text}} + \tfrac{1}{2} \bar{f}_{\mathrm{vision}}$, balancing text and all-vision tokens.

\noindent\textbf{(c) Text+Key:}\quad $s = \tfrac{1}{2} \bar{f}_{\mathrm{text}} + \tfrac{1}{2} \bar{f}_{\mathrm{key}}$, replacing all vision tokens with only key vision tokens.

\noindent\textbf{(d) Text+Redundant:}\quad $s = \tfrac{1}{2} \bar{f}_{\mathrm{text}} + \tfrac{1}{2} \bar{f}_{\mathrm{red}}$, replacing vision with only redundant vision tokens.

Under each strategy, experts with the highest scores are promoted first. Comparing (a) vs.\ (b) isolates the effect of modality-balanced ranking (inter-modality bias), while comparing (c) and (d) against (b) isolates the effect of which vision tokens drive the ranking (intra-vision bias).

Figure~\ref{fig:concept-verify} reports the accuracy curves on ChartQA and MMBench. On both benchmarks, (b) substantially outperforms (a) at every protection level, confirming the importance of separately accounting for text and vision expert preferences. (c) further improves upon (b), while (d) falls slightly below, showing that key vision tokens provide a more faithful signal of vision-side expert importance than redundant ones. Together, these comparisons validate both dimensions of frequency bias.

\section{Method}
\label{sec:method}


The concept verification in \S\ref{sec:concept_verification} shows that modality-aware frequency ranking significantly outperforms the global frequency baseline in expert importance estimation. However, frequency reflects \textit{how often} an expert is selected, which does not fully align with \textit{how much} its quantization degrades model performance. As evidenced by the outliers in Figure~\ref{fig:cross-modal-bias}(b), experts with similar frequencies can exhibit vastly different quantization sensitivities; moreover, even for equally important experts, different bit-widths yield different precision--efficiency trade-offs. To address these, we build upon modality-wise frequency and further introduce modality-wise quantization sensitivity to form a more comprehensive importance metric, then formulate the bit-width allocation as an integer linear programming (ILP) problem that jointly optimizes all expert assignments under a given bit budget. The overall pipeline is illustrated in Figure~\ref{fig:pipeline}.

\paragraph{Step 1: Modality-wise Expert Selection Frequency.}
Following the analysis in \S\ref{sec:motivation}, we collect expert selection frequencies separately for text tokens and key vision tokens (the top 20\% of vision tokens with the highest cumulative attention toward text tokens in each layer) on the calibration set. For each MoE layer, we count each expert's selections over all text tokens and all key vision tokens, respectively, and normalize the counts layer-wise so that each modality's distribution sums to one within every layer. Collecting these normalized frequencies across all layers gives $\bar{f}_t(e)$ and $\bar{f}_v(e)$ for every expert $e$ in the model.

\paragraph{Step 2: Modality-wise Quantization Sensitivity.}
While frequency reflects which experts are important, quantization sensitivity measures how much each expert's quantization hurts model outputs. For each expert $e$ and candidate bit-width $b$, we quantize \textit{only} expert $e$ to $b$ bits while keeping all other parameters at full precision, and run a forward pass on the calibration set. To disentangle per-modality effects, the quantized weights are applied exclusively to one modality at a time: for text-side sensitivity $\delta_T(e, b)$, only text tokens routed to expert $e$ use the quantized weights; for vision-side sensitivity $\delta_V(e, b)$, only key vision tokens do. Sensitivity is quantified as the KL divergence between the output logits of the original and partially quantized models, averaged over the calibration set:
\begin{equation}
    \delta_M(e, b) = \frac{1}{|\mathcal{D}|} \sum_{x \in \mathcal{D}} D_{\mathrm{KL}}\!\Big( p(x) \;\Big\|\; p^{(e \to b)}_M(x) \Big),
\end{equation}
where $M \in \{T, V\}$ denotes the modality, $p(x)$ the full-precision output logits on sample $x$, and $p^{(e \to b)}_M(x)$ the logits when expert $e$ is quantized to $b$ bits with the quantized weights applied only to modality-$M$ tokens. This per-expert, per-modality isolation ensures each sensitivity score faithfully reflects the impact of quantizing expert $e$ on the corresponding modality.


\paragraph{Step 3: Bit-width Allocation via ILP.}
Equipped with modality-wise frequencies and sensitivities, we formulate expert-level bit-width allocation as an ILP problem, a formulation also adopted by prior MoE quantization work but with modality-agnostic importance terms~\citep{huang2025mixturecompressormixtureofexpertsllms,duanmu2025mxmoemixedprecisionquantizationmoe}. The resulting ILP can be solved within seconds for the entire model. We introduce a binary indicator $x_{e,b} \in \{0,1\}$ for each expert $e$ and candidate bit-width $b \in \mathcal{B} = \{1,2,3,4\}$, where $x_{e,b}=1$ iff expert $e$ is quantized to $b$ bits. The ILP is formulated as:
\begin{equation}
    \min \sum_{e} \sum_{b \in \mathcal{B}} \Big[ \bar{f}_t(e) \cdot \delta_T(e, b) + \bar{f}_v(e) \cdot \delta_V(e, b) \Big] \, x_{e,b},
    \label{eq:objective}
\end{equation}
\begin{equation}
    \textsc{s.t.} \;\; \sum_{e} \sum_{b \in \mathcal{B}} b \cdot x_{e,b} = n \cdot k, \quad \sum_{b \in \mathcal{B}} x_{e,b} = 1, \;\; \forall e,
    \label{eq:constraint}
\end{equation}
where $n$ is the number of experts and $k$ is the target average bit-width. Each term $\bar{f}_M(e) \cdot \delta_M(e, b)$ couples \textit{how critical} expert $e$ is to modality $M$ with \textit{how much} it degrades that modality at bit-width $b$, and summing over $M \in \{T, V\}$ yields a modality-balanced cost. Minimizing this cost under the budget constraint naturally drives the solver to assign higher precision to experts that are both frequently routed and highly sensitive, while compressing the rest aggressively.


\section{Experiments}


In this section, we first describe the experimental setup, then present comprehensive results demonstrating MODE outperforms existing methods across different low-bit quantization settings when applied to MoE-MLLMs. Finally, we show MODE is orthogonal to the recently popular rotation-based quantization techniques and can be combined with them to further reduce quantization loss.

\subsection{Experimental Setup}

\paragraph{Models and Benchmarks.}
We conduct experiments on Qwen3-VL-30B-A3B-Instruct, Kimi-VL-A3B-Instruct, and InternVL3.5-30B-A3B~\citep{wang2025internvl3_5} to cover diverse MoE-MLLMs architectures and validate the generalizability of our approach. Results for InternVL3.5-30B-A3B are provided in Appendix~\ref{app:internvl} due to space constraints.

We evaluate on multiple multimodal benchmarks covering diverse capability dimensions, including ChartQA~\citep{masry-etal-2022-chartqa}, InfoVQA~\citep{DBLP:journals/corr/abs-2104-12756}, and TextVQA~\citep{singh2019towards} for OCR-oriented visual question answering; VizWiz-VQA~\citep{gurari2018vizwiz}, MME-RealWorld~\citep{zhang2025mmerealworldmultimodalllmchallenge}, and GQA~\citep{hudson2019gqa} for real-world visual understanding; MMMU~\citep{yue2023mmmu}, MMStar~\citep{chen2024we}, and MMBench~\citep{liu2024mmbench} for general multimodal reasoning; and POPE~\citep{li2023evaluating} for object hallucination.



\paragraph{Implementation Details.}
We perform weight-only quantization and report results at average bit-widths of 3-bit and 2-bit. 
Since the attention and router modules account for less than 3\% of the total LLM parameters yet significantly impact overall accuracy, we fix them at 4-bit and adjust the expert bit-width budget accordingly to maintain the target average bit-width; we provide detailed justification and ablation in Appendix~\ref{app:attn-protection}.
In the main experiments, we do not quantize the vision encoder, as its parameter count is small relative to the LLM backbone and has negligible impact on the overall memory footprint, which also aligns with most prior MoE-MLLM quantization work. Nevertheless, we additionally report results with the vision encoder quantized in Appendix~\ref{app:quant-vision} for reference.

After the mixed-precision bit-widths are determined, we apply GPTQ for weight calibration. The calibration set consists of 512 text-image pairs from ShareGPT4V, which is the same set used for collecting expert routing frequencies. We adopt symmetric quantization with a group size of 128. All evaluations are conducted using the open-source \texttt{lmms-eval}~\citep{lmms_eval2024} framework.

\begin{table*}[t]
\centering
\caption{Main quantization results on Qwen3-VL-30B-A3B-Instruct and Kimi-VL-A3B-Instruct~\citep{kimiteam2025kimivltechnicalreport} across 10 multimodal benchmarks. The best results under each setting are shown in \textbf{bold}.}
\label{tab:main}
\resizebox{\textwidth}{!}{
\begin{tabular}{ll l cccccccccc c}
\toprule
\multirow{2}{*}{Model} & \multirow{2}{*}{Avg-Bit} & \multirow{2}{*}{Method} 
  & \multicolumn{10}{c}{Benchmarks} & \multirow{2}{*}{Avg.} \\
\cmidrule(lr){4-13}
& & & MMMU & MMB & MMStar & MME-R & ChartQA & TextV & InfoV & GQA & VizWiz & POPE & \\
\midrule

\multirow{16}{*}{\makecell{\includegraphics[height=1.2em]{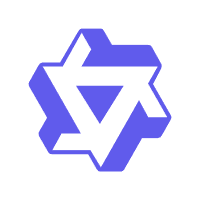} \\ Qwen3-VL\\-30B-A3B\\-Instruct}}
  & \textbf{BF16} & -- 
    & 52.56 & 86.86 & 60.05 & 52.48 & 85.20 & 83.37 & 81.46 & 62.96 & 71.64 & 89.92 & 72.65 \\
\cmidrule(lr){2-14}
  & \multirow{7}{*}{W3}
  &  GPTQ     & 44.00 & 76.63 & 45.75 & 45.29 & 66.16 & 76.58 & 66.23 & 57.73 & 62.63 & 88.79 & 62.98 \\
  & & MBQ      & 45.08 & 77.24 & 47.48 & 45.71 & 68.16 & 77.98 & 68.42 & 58.44 & 64.52 & 88.85 & 64.19 \\
  & & Speed-Q  & 46.11 & 78.65 & 49.21 & 46.46 & 73.20 & 78.25 & 71.37 & 58.92 & 65.11 & 88.82 & 65.61 \\
  & & MC-MoE   & 47.56 & 80.01 & 52.01 & 47.88 & 79.94 & 77.93 & 73.65 & 57.30 & 65.55 & 89.21 & 67.10 \\
  & & MoEQuant & 46.05 & 78.21 & 48.74 & 46.03 & 71.40 & 77.05 & 70.81 & 58.72 & 64.71 & 88.69 & 65.04 \\
  & & VEQ-ME   & 48.04 & 80.15 & 52.58 & 47.41 & 81.12 & 78.58 & 73.23 & 58.19 & 66.02 & 88.96 & 67.43 \\
  & & \cellcolor{green!10}\textbf{MODE (ours)} 
    & \cellcolor{green!10}\textbf{50.33} & \cellcolor{green!10}\textbf{82.56} & \cellcolor{green!10}\textbf{55.68} & \cellcolor{green!10}\textbf{50.77} 
    & \cellcolor{green!10}\textbf{83.00} & \cellcolor{green!10}\textbf{80.65} & \cellcolor{green!10}\textbf{76.62} & \cellcolor{green!10}\textbf{60.21} 
    & \cellcolor{green!10}\textbf{68.82} & \cellcolor{green!10}\textbf{89.44} & \cellcolor{green!10}\textbf{69.81} {\scriptsize(+2.38)} \\
\cmidrule(lr){2-14}
  & \multirow{7}{*}{W2}
  &  GPTQ     & 25.56 & 22.46 & 30.48 & 12.94 & 31.12 & 40.98 & 39.07 & 36.89 & 37.55 & 50.84 & 32.79 \\
  & & MBQ      & 35.71 & 61.23 & 36.27 & 31.63 & 57.44 & 59.84 & 53.19 & 47.16 & 48.78 & 72.37 & 50.36 \\
  & & Speed-Q  & 36.92 & 63.67 & 37.55 & 33.98 & 59.23 & 62.63 & 53.42 & 46.98 & 50.13 & 75.41 & 51.99 \\
  & & MC-MoE   & 39.78 & 66.74 & 40.48 & 40.48 & 63.12 & 70.42 & 56.74 & 53.10 & 55.32 & 80.86 & 56.70 \\
  & & MoEQuant & 36.98 & 64.49 & 38.13 & 38.31 & 60.18 & 61.32 & 56.07 & 48.01 & 48.66 & 73.53 & 52.57 \\
  & & VEQ-ME   & 41.11 & 71.68 & 42.08 & 38.97 & 67.16 & 72.38 & 59.07 & 54.89 & 59.95 & 85.56 & 59.29 \\
  & & \cellcolor{green!10}\textbf{MODE (ours)} 
    & \cellcolor{green!10}\textbf{44.63} & \cellcolor{green!10}\textbf{75.71} & \cellcolor{green!10}\textbf{45.34} & \cellcolor{green!10}\textbf{44.85} 
    & \cellcolor{green!10}\textbf{73.11} & \cellcolor{green!10}\textbf{75.45} & \cellcolor{green!10}\textbf{66.19} & \cellcolor{green!10}\textbf{56.99} 
    & \cellcolor{green!10}\textbf{63.39} & \cellcolor{green!10}\textbf{87.28} & \cellcolor{green!10}\textbf{63.29} {\scriptsize(+4.00)} \\
\midrule

\multirow{16}{*}{\makecell{\includegraphics[height=1.2em]{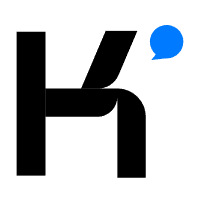} \\ Kimi-VL\\-A3B\\-Instruct}}
  & \textbf{BF16} & -- 
    & 52.33 & 82.99 & 49.41 & 44.45 & 89.44 & 88.69 & 83.61 & 62.72 & 70.59 & 87.17 & 71.14 \\
\cmidrule(lr){2-14}
  & \multirow{7}{*}{W3}
  &  GPTQ     & 45.67 & 75.04 & 43.14 & 37.27 & 82.04 & 82.05 & 69.51 & 56.05 & 62.07 & 86.06 & 63.89 \\
  & & MBQ      & 46.02 & 76.21 & 44.88 & 38.52 & 83.13 & 81.73 & 72.69 & 56.86 & 63.91 & 86.73 & 65.07 \\
  & & Speed-Q  & 45.99 & 76.55 & 45.57 & 39.21 & 82.97 & 83.01 & 74.27 & 57.33 & 63.89 & 86.32 & 65.51 \\
  & & MC-MoE   & 46.67 & 78.13 & 46.14 & 40.02 & 84.26 & 85.06 & 75.53 & 59.05 & 64.42 & 86.53 & 66.58 \\
  & & MoEQuant & 46.31 & 76.39 & 45.22 & 39.45 & 83.77 & 82.93 & 74.33 & 56.93 & 63.56 & 86.84 & 65.57 \\
  & & VEQ-ME   & 47.00 & 77.84 & 46.82 & 39.97 & 85.64 & 84.24 & 77.24 & 58.17 & 66.39 & 86.72 & 67.00 \\
  & & \cellcolor{green!10}\textbf{MODE (ours)} 
    & \cellcolor{green!10}\textbf{48.51} & \cellcolor{green!10}\textbf{80.84} & \cellcolor{green!10}\textbf{48.66} & \cellcolor{green!10}\textbf{41.59} 
    & \cellcolor{green!10}\textbf{87.26} & \cellcolor{green!10}\textbf{86.24} & \cellcolor{green!10}\textbf{79.87} & \cellcolor{green!10}\textbf{61.11} 
    & \cellcolor{green!10}\textbf{69.42} & \cellcolor{green!10}\textbf{87.10} & \cellcolor{green!10}\textbf{69.06} {\scriptsize(+2.06)} \\
\cmidrule(lr){2-14}
  & \multirow{7}{*}{W2}
  &  GPTQ     & 27.33 & 23.44 & 23.01 & 24.23 & 27.62 & 48.02 & 22.42 & 37.09 & 25.20 & 70.29 & 32.87 \\
  & & MBQ      & 31.48 & 38.21 & 33.54 & 29.57 & 36.77 & 59.98 & 45.17 & 41.79 & 39.33 & 74.31 & 43.02 \\
  & & Speed-Q  & 35.33 & 51.44 & 36.01 & 31.98 & 44.62 & 62.02 & 50.42 & 47.09 & 49.20 & 77.29 & 48.54 \\
  & & MC-MoE   & 40.92 & 72.61 & 41.52 & 36.01 & 66.49 & 75.15 & 64.23 & 55.97 & 58.26 & 82.58 & 59.37 \\
  & & MoEQuant & 37.63 & 63.16 & 35.78 & 33.12 & 53.24 & 68.84 & 55.51 & 50.96 & 53.69 & 79.13 & 53.11 \\
  & & VEQ-ME   & 39.78 & 72.16 & 42.13 & 35.52 & 62.24 & 74.0 6 & 64.94 & 55.45 & 57.04 & 83.09 & 58.64 \\
  & & \cellcolor{green!10}\textbf{MODE (ours)} 
    & \cellcolor{green!10}\textbf{45.11} & \cellcolor{green!10}\textbf{75.23} & \cellcolor{green!10}\textbf{44.11} & \cellcolor{green!10}\textbf{38.71} 
    & \cellcolor{green!10}\textbf{75.20} & \cellcolor{green!10}\textbf{79.38} & \cellcolor{green!10}\textbf{67.75} & \cellcolor{green!10}\textbf{58.30} 
    & \cellcolor{green!10}\textbf{64.52} & \cellcolor{green!10}\textbf{85.46} & \cellcolor{green!10}\textbf{63.38} {\scriptsize(+4.01)} \\
\bottomrule
\end{tabular}
}
\end{table*}

\subsection{Main Results}


\paragraph{Baselines.}
To provide a thorough and convincing comparison, besides the basic uniform quantization method GPTQ~\citep{frantar2023gptqaccurateposttrainingquantization}, we cover three categories of PTQ methods designed for MLLMs, MoE-LLMs, and MoE-MLLMs, respectively:
\begin{itemize}[leftmargin=*,itemsep=1pt,topsep=2pt]
    \item \textbf{For MLLMs:} MBQ~\citep{li2024mbq} and Speed-Q~\citep{guo2025speedq}, both of which account for inter-modality differences during quantization.
    \item \textbf{For MoE-LLMs:} MC-MoE and MoEQuant~\citep{hu2025moequantenhancingquantizationmixtureofexperts}, which are tailored for MoE architectures, where MC-MoE similarly employs mixed-precision quantization across experts.
    \item \textbf{For MoE-MLLMs:} VEQ-ME~\citep{qin2026veqmodalityadaptivequantizationmoe}, the most recent PTQ method explicitly designed for MoE-MLLMs based on GPTQ.
\end{itemize}
All baseline results are reproduced using their official code repositories. For a fairer comparison with VEQ-ME, whose originally reported results suffer over 10\% average accuracy degradation at 3-bit due to unprotected attention weights, we reproduce all its results with the same protection applied.



\paragraph{Results.}
Table~\ref{tab:main} summarizes the quantization results on two representative MoE-MLLMs across 10 multimodal benchmarks under 3-bit and 2-bit settings. MLLM-oriented methods such as MBQ and Speed-Q consider modality-level differences but do not explicitly model the structural characteristics of MoE, which leads to noticeable performance degradation. 
This becomes more evident under 2-bit quantization, where the average accuracy drops by over 20\%.
MoE-targeted methods like MC-MoE mitigate severe collapse through mixed-precision bit allocation across experts, yet still incur a non-trivial quantization loss, as they overlook the modality imbalance in expert usage inherent to MoE-MLLMs. VEQ-ME achieves better results than both categories in most settings, though the improvements remain relatively limited.

Our method jointly addresses the heterogeneous expert contributions inherent to MoE architecture and the inter-modality bias together with intra-visual redundancy characteristic of MLLMs, yielding consistent improvements across both models and bit-width settings. At 3-bit, the average accuracy loss is kept within 2.9\% and 2.1\% on Qwen3-VL and Kimi-VL respectively, substantially improving practical usability. At 2-bit, our method outperforms the best competing approach by over 4\% on average, effectively narrowing the performance gap under extreme low-bit quantization.

\subsection{Compatibility with Rotation-based Quantization}

Rotation-based quantization, exemplified by QuaRot~\citep{NEURIPS2024_b5b93943}, has become a mainstream PTQ technique that applies random Hadamard rotations to suppress outliers in weights and activations, thereby reducing quantization error at its source. Since our method focuses on mixed-precision bit allocation along the expert dimension, it is orthogonal to rotation-based approaches. We therefore examine whether combining QuaRot on top of our method can yield further gains.

\begin{table}[t]
\centering
\caption{Effect of combining rotation-based quantization (QuaRot) with our method on Qwen3-VL-30B-A3B-Instruct. ``+Rot'' denotes applying Hadamard rotation on top of our mixed-precision framework.}
\label{tab:rotation}
\resizebox{\linewidth}{!}{
\begin{tabular}{l|c|ccc}
\toprule
\textbf{Method} & \textbf{Bits} & \textbf{MMMU} & \textbf{VizWiz} & \textbf{TextVQA} \\
\midrule
Baseline (BF16) & 16 & 52.56 & 71.64 & 83.37 \\
\midrule
Ours             & 3 & 50.33 & 68.82 & 80.65 \\
\rowcolor{blue!5} Ours (+Rot)      & 3 & \textbf{50.81} & \textbf{69.17} & \textbf{81.50} \\
\midrule
Ours             & 2 & 44.63 & 63.39 & 75.45 \\
\rowcolor{blue!5} Ours (+Rot)      & 2 & \textbf{46.49} & \textbf{66.02} & \textbf{76.94} \\
\bottomrule
\end{tabular}
}
\vskip -0.1in
\end{table}

As shown in Table~\ref{tab:rotation}, integrating QuaRot consistently improves performance on top of our method. The benefit is particularly pronounced under the extreme 2-bit setting, where rotation yields an average improvement of around 1.5\%, confirming the two techniques address complementary sources of quantization error and can be effectively combined.

\subsection{Robustness to Calibration Set}

Our method collects expert activation frequencies under different modalities on a calibration set as one indicator of expert importance. It is therefore necessary to verify that this signal is robust to the choice of calibration data. We run our full pipeline with three different calibration sets---ShareGPT4V, Flickr30k~\citep{young2014image}, and LLaVA-Next~\citep{liu2024llavanext}, and evaluate the resulting quantized models on the same downstream benchmarks. As shown in Table~\ref{tab:calib_acc}, switching the calibration set introduces negligible variation in average accuracy, 
confirming that our method is not sensitive to the specific calibration data used. The underlying reason is that MoE-based MLLMs exhibit similar expert selection preferences across different datasets, as demonstrated in Section~\ref{sec:intra-vision-bias} where key vision tokens converge to a stable set of preferred experts regardless of data composition. 

\begin{table}[hbt]
\centering
\caption{Performance of our method at 3-bit on Qwen3-VL-30B-A3B-Instruct with different calibration sets.}
\vskip -0.1in
\label{tab:calib_acc}
\resizebox{\linewidth}{!}{
\begin{tabular}{l|ccccc|c}
\toprule
\textbf{Calib. Set} & \textbf{MMMU} & \textbf{VizWiz} & \textbf{InfoVQA} & \textbf{MMStar} & \textbf{ChartQA} & \textbf{Avg.} \\
\midrule
Baseline (BF16) & 52.56 & 71.64 & 81.46 & 60.05 & 85.20 & 70.18 \\
\midrule
ShareGPT4V & 50.33 & 68.82 & 76.62 & 55.68 & 83.00 & 66.89 \\
Flickr30k  & 49.61 & 68.27 & 77.04 & 54.75 & 82.54 & 66.44 \\
LLaVA-Next & 50.78 & 68.91 & 76.56 & 56.02 & 83.36 & 67.13 \\
\bottomrule
\end{tabular}
}
\end{table}

\section{Deployment Efficiency}
\label{sec:efficiency}
\begin{figure}[tbh]
    \centering
    \includegraphics[width=\linewidth]{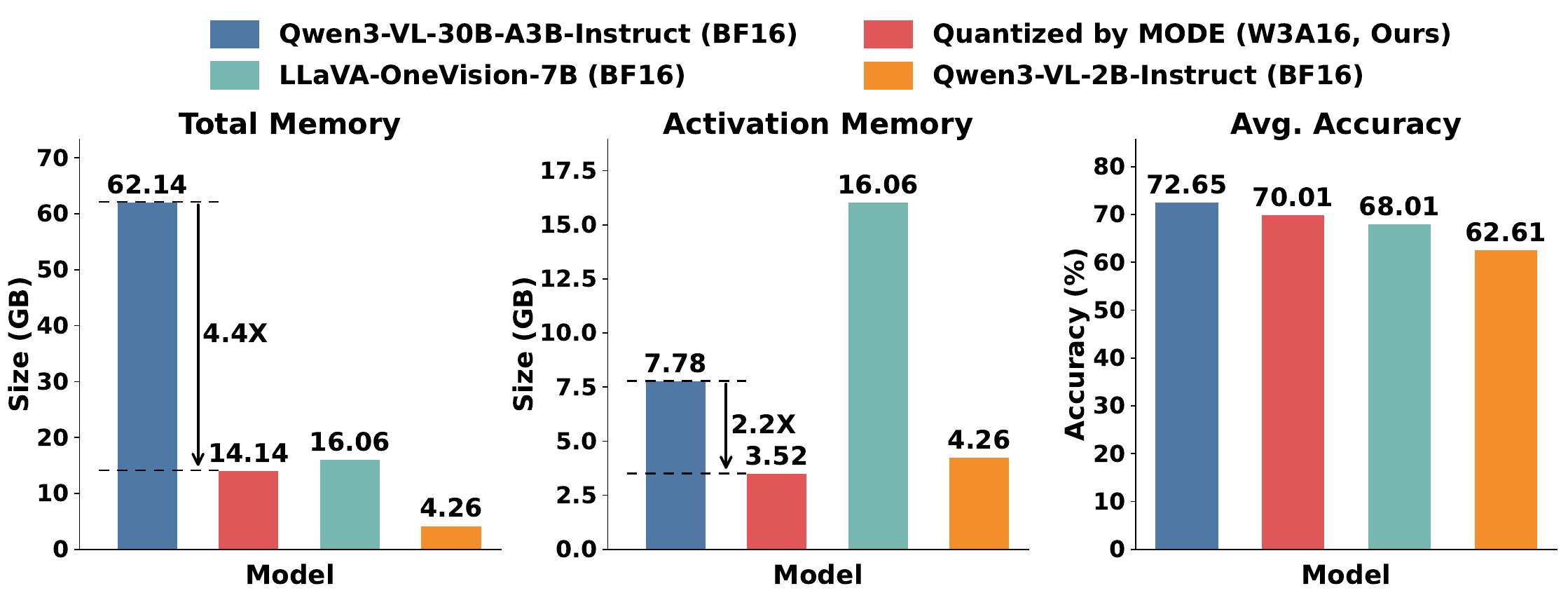}
    \vskip -0.1in
    \caption{Deployment efficiency of Qwen3-VL-30B-A3B-Instruct quantized by MODE (W3A16) in terms of total / activation memory, and accuracy retention.}
    \label{fig:efficiency}
    \vskip -0.1in
\end{figure}

For real quantization, we use BitBLAS~\citep{ladder-osdi24} to store quantized weights at varying bit-widths and perform mixed-precision BLAS operations on GPUs. To demonstrate efficiency, we take Qwen3-VL-30B-A3B-Instruct quantized to an average of 3 bits as an example. Figure~\ref{fig:efficiency} reports its total weight memory, activation memory, and average accuracy across ten benchmarks, along with two dense MLLMs as references: LLaVA-OneVision-7B~\citep{li2024llavaonevisioneasyvisualtask}, which has a comparable total weight memory, and Qwen3-VL-2B-Instruct, which has a comparable activation memory.

After 3-bit quantization, the total weight memory drops from 62\,GB to around 14\,GB, fitting comfortably within a single consumer-grade GPU such as RTX 4090 (24\,GB). Compared with LLaVA-OneVision-7B at a similar total weight memory, the quantized model requires far less activation memory thanks to the sparse activation nature of MoE, while achieving higher accuracy. Compared with Qwen3-VL-2B-Instruct at a comparable activation memory, the quantized model surpasses it by a large margin in accuracy. These results confirm that our method enables practical deployment of large-scale MoE-MLLMs on resource-constrained hardware with only a minor performance drop.

\begin{table*}[t]
\centering
\small
\setlength{\tabcolsep}{10pt}
\renewcommand{\arraystretch}{1.2}
\caption{End-to-end (E2E) latency and prefill/decode throughput on Qwen3-VL-30B-A3B-Instruct with a maximum output length of 64. BF16 is deployed on two GPUs via Hugging Face Transformers due to its memory requirement, while MODE-W3A16/W2A16 runs on a single GPU via BitBLAS. \textcolor{gray}{Gray} numbers denote the speedup over the BF16 baseline at the same batch size (BS).}
\label{tab:latency}
\newcommand{\up}[1]{\,\textcolor{gray}{\scriptsize$#1\times$}}
\begin{tabular}{llcccc}
\toprule
Precision & BS & E2E (s) & Output (tok/s) & Prefill (tok/s) & Decode (tok/s) \\
\midrule
\multirow{3}{*}{BF16}
 & 1 & 10.07 & 6.36 & 34.37 & 16.27 \\
 & 2 & 18.65 & 6.87 & 37.31 & 28.73 \\
 & 4 & 34.67 & 7.39 & 36.67 & 55.93 \\
\midrule
\rowcolor{gray!8}
\cellcolor{white}
 & 1 & 7.82\up{1.29}  & 8.19\up{1.29} & 42.96\up{1.25} & 22.45\up{1.38} \\
\rowcolor{gray!8}
\cellcolor{white}
 & 2 & 14.55\up{1.28} & 8.80\up{1.28} & 47.38\up{1.27} & 40.51\up{1.41} \\
\rowcolor{gray!8}
\cellcolor{white}\multirow{-3}{*}{W3A16}
 & 4 & 27.28\up{1.27} & 9.39\up{1.27} & 46.94\up{1.28} & 79.42\up{1.42} \\
\midrule
\rowcolor{blue!6}
\cellcolor{white}
 & 1 & 7.35\up{1.37}  & 8.71\up{1.37} & 45.71\up{1.33} & 23.92\up{1.47} \\
\rowcolor{blue!6}
\cellcolor{white}
 & 2 & 13.75\up{1.36} & 9.31\up{1.36} & 50.37\up{1.35} & 42.81\up{1.49} \\
\rowcolor{blue!6}
\cellcolor{white}\multirow{-3}{*}{W2A16}
 & 4 & 25.80\up{1.34} & 9.92\up{1.34} & 50.24\up{1.37} & 86.13\up{1.54} \\
\bottomrule
\end{tabular}
\end{table*}

\paragraph{Inference speed.}
Beyond memory footprint, we further examine whether the memory savings translate into practical inference speedups. We randomly sample inputs from ShareGPT4V and measure the end-to-end (E2E) latency together with the prefill and decode throughput on Qwen3-VL-30B-A3B-Instruct, under batch sizes of $1$, $2$, and $4$ with a maximum output length of $64$. Since the BF16 model exceeds the memory capacity of a single GPU, it is deployed with Hugging Face Transformers across two GPUs, whereas MODE-W3A16 and MODE-W2A16 run with BitBLAS on a single GPU. As reported in Table~\ref{tab:latency}, MODE consistently reduces E2E latency and improves both prefill and decode throughput across all batch sizes. At the 3-bit setting, MODE achieves up to a $1.29\times$ E2E speedup and a $1.42\times$ decode throughput gain, and the 2-bit setting further pushes these to $1.37\times$ and $1.54\times$, respectively. Notably, these gains are obtained while using only a single GPU rather than the two GPUs required by BF16, confirming that the memory savings of MODE directly lead to tangible deployment speedups on resource-constrained hardware.



\section{Conclusion}
\label{sec:conclusion}

We propose MODE, a mixed-precision expert quantization method for MoE-MLLMs. Unlike prior work allocating bit-widths based solely on expert activation statistics, MODE explicitly accounts for both the \textit{inter-modal bias} between textual and visual tokens and the \textit{intra-visual-modal bias} across different visual semantics, enabling a more faithful characterization of expert importance in the multimodal setting. Under a fixed bit budget, MODE efficiently allocates precision across experts, 
substantially mitigating the performance degradation of weight-only quantization under low-bit regimes. 
Extensive experiments demonstrate that MODE consistently outperforms existing baselines, narrowing the gap toward practical deployment of MoE-MLLMs on resource-constrained devices.








\section*{Limitations}
While our work demonstrates the effectiveness of MODE for quantizing MoE-MLLMs, we acknowledge the following limitations and directions for future improvement.

\paragraph{Scaling to larger MoE-MLLMs.} Our experiments are conducted on Qwen3-VL-30B-A3B-Instruct, Kimi-VL-A3B-Instruct, and InternVL3.5-30B-A3B, which are deliberately chosen to cover a diverse set of model architectures and thereby validate the generalization of our method. However, due to computational resource constraints, we have not yet verified MODE on larger-scale MoE-MLLMs such as Qwen3-VL-235B-A22B. We leave the evaluation on such larger models to future work when more computational resources become available.

\paragraph{Quantization of the vision module.} As discussed in Appendix~\ref{app:quant-vision}, our method is primarily designed for the LLM part of MoE-MLLMs and does not incorporate a dedicated design for the vision module. Although the vision module occupies only a small fraction of the total memory footprint and thus contributes marginally to the deployment cost, we leave a thorough investigation of vision-module quantization in MoE-MLLMs to future work in pursuit of a more comprehensive PTQ design.


\paragraph{Deployment efficiency in modern inference frameworks.} As reported in Section~\ref{sec:efficiency}, for real quantization we leverage BitBLAS~\citep{ladder-osdi24} to store quantized weights at varying bit-widths and perform mixed-precision BLAS operations on GPUs, which achieves significant reductions in memory consumption and, as validated on the Hugging Face Transformers backbone, translates into consistent end-to-end latency reductions and prefill/decode throughput improvements over the full-precision baseline. Nevertheless, we honestly note that these speedups are obtained relative to a Transformers-based baseline, and our current implementation has not yet been adapted to highly optimized inference frameworks such as vLLM~\citep{kwon2023efficientmemorymanagementlarge}. This is mainly because vLLM and similar frameworks incorporate extensive system-level optimizations tailored to MoE architectures, including efficient KV cache management and fused kernels for MoE layers, while currently vLLM does not natively support integrating mixed-precision quantization into its runtime. As a result, the absolute inference speed of the quantized model still lags behind full-precision models served by such frameworks. We hope to integrate our mixed-precision expert quantization scheme into high-performance inference frameworks like vLLM in the future, so as to translate the memory savings into larger practical speedups.

\section*{Ethical Considerations}
\label{sec:ethics}

This work focuses on post-training quantization for MoE-based Multimodal Large Language Models, which is a model compression technique and does not involve the collection of new data or human subjects. All models and benchmarks used in our experiments are publicly available and employed in accordance with their respective licenses and intended research use. Since our method reduces the memory and hardware requirements for deploying MoE-MLLMs, it may contribute positively by lowering the barrier to accessing such models and reducing energy consumption during inference.

\section*{Acknowledgments}
This work was supported by the National Natural Science Foundation of China (Grant No. 62572471, 62531026, 62437001), the Zhongguancun Academy (Grant No. C20250505), the Natural Science Foundation of Jiangsu Province under Grant BK20243051.


\bibliography{custom}

\appendix


\section{Complete Per-Layer Heatmap of Intra-Vision Expert Frequency Bias}
\label{sec:appendix-intra-vision-full}

\begin{figure*}[t]
    \centering
    \includegraphics[width=0.99\linewidth]{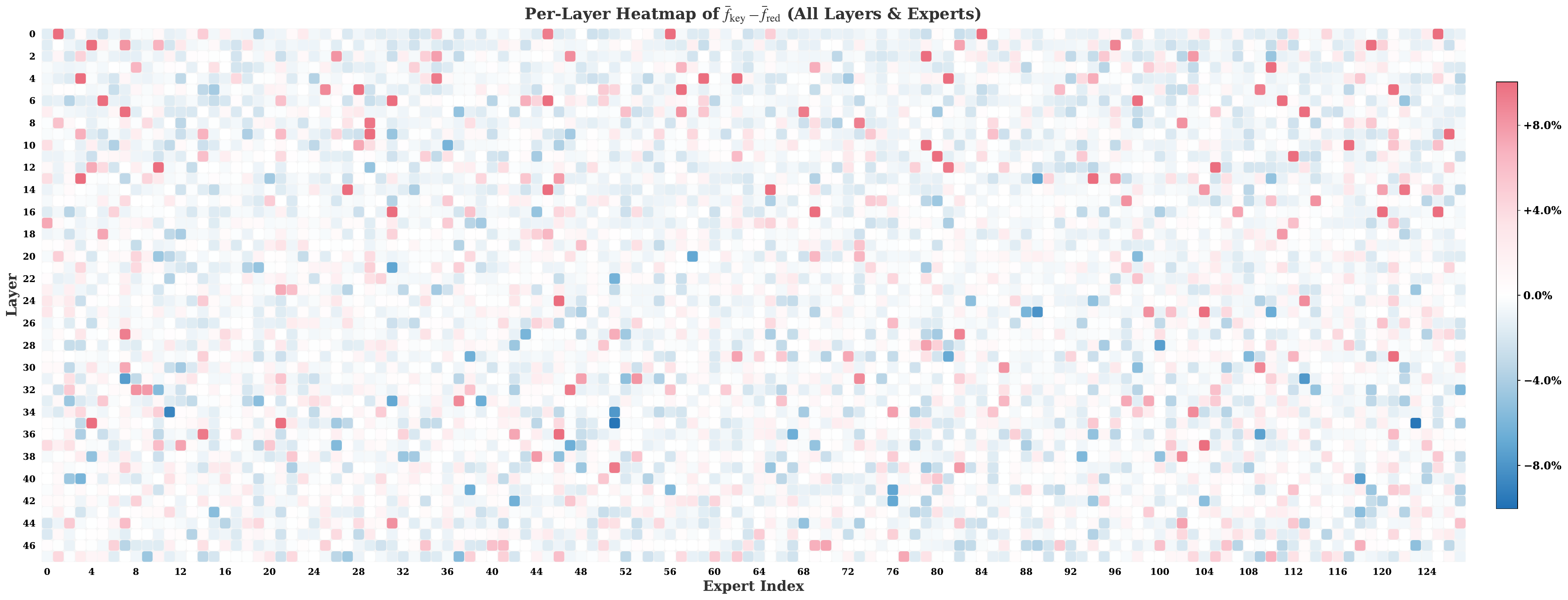}
    \caption{Complete per-layer heatmap of $\bar{f}_{\mathrm{key}} - \bar{f}_{\mathrm{red}}$ on Qwen3-VL-30B-A3B-Instruct, covering all 48 layers and all 128 experts per layer.}
    \label{fig:intra-vision-bias-full}
\end{figure*}

In Section~\ref{sec:intra-vision-bias}, due to space constraints, Figure~\ref{fig:intra-vision-bias}(b) only visualizes $\bar{f}_{\mathrm{key}} - \bar{f}_{\mathrm{red}}$ for the first 16 layers and the first 48 experts. However, Qwen3-VL-30B-A3B-Instruct contains 48 layers in total, with 128 experts per layer. For completeness, we provide the full heatmap covering all layers and all experts in Figure~\ref{fig:intra-vision-bias-full}.

\paragraph{On the magnitude of frequency differences.}
A natural question is why the absolute values of $\bar{f}_{\mathrm{key}} - \bar{f}_{\mathrm{red}}$ in Figure~\ref{fig:intra-vision-bias-full} (and in Figure~\ref{fig:intra-vision-bias}(b)) appear noticeably smaller than the routing frequencies reported in Figure~\ref{fig:cross-modal-bias} of Section~\ref{sec:cross-modal-bias}, where the maximum value can exceed $25\%$. The discrepancy stems from a difference in normalization rather than a difference in the underlying routing behavior.

In Qwen3-VL-30B-A3B-Instruct, each MoE layer contains $128$ experts, and each token activates the top-$8$ experts. Consequently, even under a perfectly uniform routing distribution, the expected selection frequency of any single expert is
\[
\frac{8}{128} = 6.25\%.
\]
The frequencies reported in Figure~\ref{fig:cross-modal-bias} are \emph{unnormalized} selection frequencies, so values around or above this $6.25\%$ baseline are expected, and peaks beyond $25\%$ indicate strong routing preferences.

In contrast, the values in Figure~\ref{fig:intra-vision-bias}(b) and Figure~\ref{fig:intra-vision-bias-full} are computed from \emph{normalized} per-token routing distributions, where the activated experts of each token sum to $1$. Under this normalization, the expected selection frequency of any single expert under uniform routing is
\[
\frac{1}{128} \approx 0.78\%.
\]
Relative to this baseline, a difference of $\bar{f}_{\mathrm{key}} - \bar{f}_{\mathrm{red}}$ on the order of $8\%$ already represents an obviously large routing-frequency gap—roughly an order of magnitude above the uniform baseline—indicating a pronounced specialization of certain experts toward dominant (key) versus redundant tokens.


\section{Additional Results on InternVL3.5-30B-A3B}
\label{app:internvl}

In the main paper, we report results on Qwen3-VL-30B-A3B-Instruct and Kimi-VL-A3B-Instruct. Due to space constraints, we defer the results on InternVL3.5-30B-A3B to this appendix. We follow the exact same evaluation protocol as in the main experiments, evaluating on the same 10 multimodal benchmarks under both average 3-bit and 2-bit settings. As shown in Table~\ref{tab:internvl}, our proposed MODE consistently outperforms both the mixed-precision MoE quantization method MC-MoE and the MoE-MLLM-tailored method VEQ-MA at both bit-widths. In particular, under the 3-bit setting, MODE limits the average accuracy degradation to within 3\% relative to the BF16 baseline (69.25 vs.\ 72.15), demonstrating the strong generalization of our method across different model families. Under the more challenging 2-bit setting, MODE still surpasses the strongest baseline VEQ-MA by about 2 points on average, further confirming its robustness under extremely low-bit quantization.

\begin{table*}[bth]
\centering
\caption{Quantization results on InternVL3.5-30B-A3B across 10 multimodal benchmarks. The best results under each setting are shown in \textbf{bold}. The number in parentheses denotes the absolute average gain of MODE over the strongest baseline.}
\label{tab:internvl}
\resizebox{\textwidth}{!}{
\begin{tabular}{ll l cccccccccc c}
\toprule
\multirow{2}{*}{Model} & \multirow{2}{*}{Avg-Bit} & \multirow{2}{*}{Method} 
  & \multicolumn{10}{c}{Benchmarks} & \multirow{2}{*}{Avg.} \\
\cmidrule(lr){4-13}
& & & MMMU & MMB & MMStar & MME-R & ChartQA & TextV & InfoV & GQA & VizWiz & POPE & \\
\midrule

\multirow{10}{*}{\makecell{\includegraphics[height=1.2em]{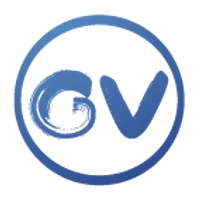} \\ InternVL3.5\\-30B-A3B}}
  & \textbf{BF16} & -- 
    & 60.11 & 86.08 & 71.25 & 49.71 & 88.12 & 78.35 & 76.27 & 62.95 & 59.23 & 89.38 & 72.15 \\
\cmidrule(lr){2-14}
  & \multirow{4}{*}{W3}
  &  GPTQ    & 52.56 & 81.19 & 63.19 & 44.17 & 77.60 & 70.34 & 65.44 & 56.01 & 45.23 & 89.13 & 64.49 \\
  & & MC-MoE  & 54.56 & 81.42 & 66.57 & 47.52 & 82.00 & 72.67 & 67.68 & 57.38 & 50.28 & 88.57 & 66.87 \\
  & & VEQ-MA  & 55.49 & 81.12 & 67.01 & 46.05 & 83.07 & 73.13 & 69.75 & 59.21 & 53.04 & 89.08 & 67.70 \\
  & & \cellcolor{green!10}\textbf{MODE (ours)} 
    & \cellcolor{green!10}\textbf{57.33} & \cellcolor{green!10}\textbf{82.56} & \cellcolor{green!10}\textbf{68.26} & \cellcolor{green!10}\textbf{47.97} 
    & \cellcolor{green!10}\textbf{85.64} & \cellcolor{green!10}\textbf{74.45} & \cellcolor{green!10}\textbf{71.53} & \cellcolor{green!10}\textbf{60.15} 
    & \cellcolor{green!10}\textbf{55.55} & \cellcolor{green!10}\textbf{89.03} & \cellcolor{green!10}\textbf{69.25} {\scriptsize(+1.55)} \\
\cmidrule(lr){2-14}
  & \multirow{4}{*}{W2}
  &  GPTQ    & 22.78 & 77.32 & 31.67 & 21.68 & 40.75 & 36.11 & 25.62 & 30.30 & 30.11 & 40.18 & 35.65 \\
  & & MC-MoE  & 44.89 & 76.43 & 57.40 & 43.79 & 70.92 & 67.94 & 60.10 & 55.63 & 44.07 & 88.18 & 60.94 \\
  & & VEQ-MA  & 46.11 & 75.76 & 59.76 & 42.33 & 72.36 & 67.55 & 61.68 & 56.04 & 48.72 & 88.46 & 61.88 \\
  & & \cellcolor{green!10}\textbf{MODE (ours)} 
    & \cellcolor{green!10}\textbf{48.67} & \cellcolor{green!10}\textbf{78.69} & \cellcolor{green!10}\textbf{60.97} & \cellcolor{green!10}\textbf{45.31} 
    & \cellcolor{green!10}\textbf{73.68} & \cellcolor{green!10}\textbf{69.83} & \cellcolor{green!10}\textbf{63.65} & \cellcolor{green!10}\textbf{58.40} 
    & \cellcolor{green!10}\textbf{50.87} & \cellcolor{green!10}\textbf{88.66} & \cellcolor{green!10}\textbf{63.87} {\scriptsize(+2.00)} \\
\bottomrule
\end{tabular}
}
\end{table*}

\section{Details of Key Vision Token Selection}
\label{app:key-vision-token}

In this appendix, we provide a detailed description of the key vision token selection used in our intra-vision expert frequency bias analysis. We first review the attention-based token importance criterion proposed by SparseVLM~\citep{zhang2024sparsevlm} and discuss why it is well-motivated for identifying redundant vision tokens (Appendix~\ref{app:sparsevlm-recap}). We then describe how we adapt this criterion to our calibration setting, with a key difference from prior pruning-oriented usage (Appendix~\ref{app:our-adaptation}). Finally, we empirically verify that the tokens selected by our procedure indeed behave as ``key'' and ``redundant'' vision tokens (Appendix~\ref{app:verification}).

\subsection{Recap of SparseVLM's Attention-based Criterion}
\label{app:sparsevlm-recap}

A growing body of work has observed that vision tokens in VLMs/MLLMs are highly redundant: a large fraction of them can be pruned with negligible degradation in downstream performance, while substantially reducing the computation and memory cost of inference~\citep{chen2024imageworth12tokens,yang2026visionziplongerbetternecessary,tan2025tokencarveinformationpreservingvisualtoken}. Among these methods, SparseVLM~\citep{zhang2024sparsevlm} stands out as a representative approach. Although many follow-up works have further improved upon it, SparseVLM remains attractive for its simplicity and effectiveness, requiring no additional modules or training. We therefore follow its core criterion in this work and adapt it to our needs.

The key idea of SparseVLM is to reuse the self-attention logits already computed inside the VLM decoder as a measure of how relevant each vision token is to the current text query. Concretely, let $\boldsymbol{A} \in \mathbb{R}^{L \times L}$ denote the self-attention matrix at a given layer, where $L$ is the total sequence length. Let $\mathbb{L}$ and $\mathbb{I}$ denote the index sets of language (text) tokens and image (vision) tokens, respectively, with $|\mathbb{L}| = L_t$ and $|\mathbb{I}| = L_v$. SparseVLM extracts the text-to-vision sub-block
\begin{equation}
\boldsymbol{P} = \boldsymbol{A}_{i,j}, \quad (i,j) \in \mathbb{L} \times \mathbb{I}, \qquad \boldsymbol{P} \in \mathbb{R}^{L_t \times L_v},
\end{equation}
and aggregates the attention each vision token receives from all text tokens:
\begin{equation}
\tilde{\boldsymbol{p}} = [\tilde{p}_1, \tilde{p}_2, \dots, \tilde{p}_{L_v}] = \frac{1}{L_t} \sum_{i=1}^{L_t} \boldsymbol{P}_i.
\end{equation}
A larger value of $\tilde{p}_j$ indicates that vision token $j$ is more relevant to the current language query and thus more important to retain. Because $\boldsymbol{A}$ is already computed during the forward pass, this estimation is essentially free in terms of additional FLOPs.

\subsection{Our Adaptation: Per-Layer Adaptive Selection}
\label{app:our-adaptation}

The most important difference between our usage and the original pruning-oriented usage in SparseVLM lies in \emph{when} and \emph{how} the selection is applied. In SparseVLM and most subsequent token-pruning methods, the decision must be made at a relatively shallow layer (e.g., the second or third decoder layer): tokens deemed redundant are physically dropped and excluded from all subsequent layers' computation, which is precisely how the speedup is achieved. However, a well-known issue with this design is that the attention distribution over vision tokens varies considerably across depths: a vision token that receives little attention in shallow layers may become highly attended in deeper layers, and vice versa. Committing to a one-shot pruning decision at a shallow layer therefore inevitably introduces selection bias.

In our pipeline, by contrast, the selection is used purely for \emph{calibration}, not for actual pruning during inference: we only need to identify which vision tokens behave as key tokens at each layer, in order to compute the modality-wise expert selection frequency. This frees us from the constraint of making an early, layer-agnostic decision. Accordingly, we apply the SparseVLM criterion \emph{independently at every layer}: at layer $\ell$, we compute $\tilde{\boldsymbol{p}}^{(\ell)}$ from that layer's attention, rank vision tokens by $\tilde{\boldsymbol{p}}^{(\ell)}$, and treat the top $20\%$ as key vision tokens and the remaining $80\%$ as redundant vision tokens. The resulting key-token set is thus adaptive to how the model attends to visual content at different depths, avoiding the cross-layer bias that hampers shallow-layer pruning.

\subsection{Empirical Verification of the Selected Key and Redundant Tokens}
\label{app:verification}

To verify that our per-layer attention-based selection indeed identifies vision tokens that are functionally ``key'' or ``redundant'' as intended, we conduct a controlled pruning experiment on Qwen3-VL-30B-A3B-Instruct using ChartQA.

\paragraph{Protocol.} At every decoder layer, we rank vision tokens by their importance score $\tilde{\boldsymbol{p}}^{(\ell)}$ as described in Appendix~\ref{app:our-adaptation}. We then prune vision tokens at increasing ratios from $5\%$ to $50\%$ in steps of $5\%$, under two opposite strategies: (i) \emph{key-first pruning}, which removes the highest-ranked tokens first, and (ii) \emph{redundant-first pruning}, which removes the lowest-ranked tokens first. We restrict the pruning ratio to at most $50\%$, making the two strategies cleanly comparable. For each pruning ratio, we evaluate the resulting model on ChartQA and plot accuracy versus pruning ratio as two curves (key-first vs.\ redundant-first).

\begin{figure}[t]
    \centering
    \includegraphics[width=\linewidth]{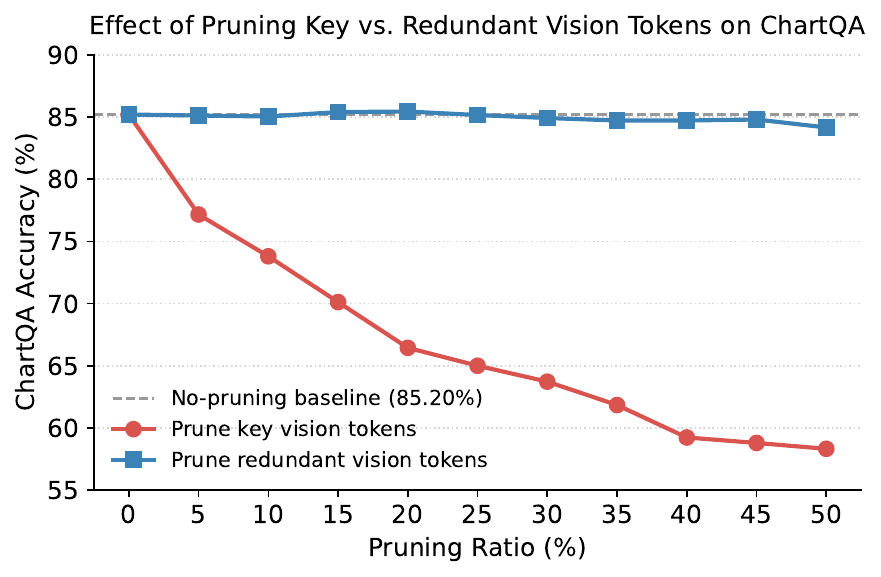}
    \caption{Effect of pruning key vs.\ redundant vision tokens on ChartQA. 
    Removing tokens identified as \emph{key} (red) causes a rapid accuracy drop, 
    whereas removing tokens identified as \emph{redundant} (blue) leaves accuracy 
    almost unchanged across the entire $5\%$--$50\%$ range.} 
    \label{fig:keyvision-verification}
\end{figure}


\paragraph{Results.} As shown in Figure~\ref{fig:keyvision-verification}, the two strategies exhibit sharply contrasting trends. Under key-first pruning, accuracy degrades sharply even at very small pruning ratios---dropping from the no-pruning baseline of $85.20\%$ to roughly $66\%$ after pruning only $20\%$ of the vision tokens---which indicates that the tokens we identify as ``key'' indeed carry information that the model genuinely relies on. As the ratio further increases, the curve flattens and decreases more gradually, since the additionally removed tokens are progressively less critical. In contrast, under redundant-first pruning, accuracy stays essentially flat throughout the entire range, fluctuating only within roughly $84\%$--$85\%$ and remaining close to the baseline even when half of the vision tokens are discarded. This pronounced asymmetry confirms that our per-layer adaptive selection reliably separates key vision tokens from redundant ones, providing a solid basis for the subsequent analysis that links key vision tokens to critical vision experts.


\section{Details of Expert Selection Frequency Collection and Cross-Dataset Similarity}
\label{app:freq-details}

In this appendix, we formally describe how expert selection frequencies are collected, normalized, and compared across calibration sets. We first introduce the general collection procedure without distinguishing token types (Appendix~\ref{app:freq-general}), then specialize it to text, key vision, and redundant vision tokens with layer-wise normalization (Appendix~\ref{app:freq-normalize}), and finally describe the cross-dataset cosine similarity (Appendix~\ref{app:freq-cosine}).

\subsection{General Expert Selection Frequency}
\label{app:freq-general}

Consider an MoE model with $L$ MoE layers and $E$ experts per layer. For each token, the router at every layer activates a top-$k$ subset of experts. Given a calibration dataset $\mathcal{D}$, we run the model on it and, at each layer $\ell$, simply count how many tokens activate each expert $e$:
\begin{equation}
c^{(\ell)}_e \;=\; \#\{\, x \in \mathcal{D} : \text{$x$ selects expert $e$ at layer $\ell$} \,\}.
\label{eq:freq-count}
\end{equation}
Stacking over $e=1,\dots,E$ gives a per-layer count vector $\boldsymbol{c}^{(\ell)} \in \mathbb{R}^E$.

\subsection{Per-Token-Type Frequency and Layer-wise Normalization}
\label{app:freq-normalize}

We further partition tokens by their role: text tokens (\textit{text}), key vision tokens (\textit{key}, top-$20\%$ identified per layer as in Appendix~\ref{app:our-adaptation}), and redundant vision tokens (\textit{red}, the remaining $80\%$). For each token type $m \in \{\textit{text},\, \textit{key},\, \textit{red}\}$, we restrict the counting in Eq.~\eqref{eq:freq-count} to tokens of type $m$ only, yielding a type-specific count $c^{(\ell)}_{m,e}$.

To make frequencies comparable across layers and across calibration sets of different sizes, we normalize $\boldsymbol{c}^{(\ell)}_{m}$ at each layer into a probability distribution over experts:
\begin{equation}
\bar{f}^{(\ell)}_{m,e} \;=\; \frac{c^{(\ell)}_{m,e}}{\sum_{e'=1}^{E} c^{(\ell)}_{m,e'}},
\label{eq:freq-norm}
\end{equation}
so that $\sum_{e} \bar{f}^{(\ell)}_{m,e} = 1$. Stacking over experts yields the normalized frequency vector $\bar{\boldsymbol{f}}^{(\ell)}_{m} \in \mathbb{R}^E$.

\paragraph{Notation across sections.} The frequency symbols used in the main text are all special cases of $\bar{\boldsymbol{f}}^{(\ell)}_{m}$, but with slightly different scopes depending on context:
\begin{itemize}[leftmargin=*]
    \item In Sec.~\ref{sec:concept_verification}, four frequencies appear: $\bar{f}_{\textit{total}}$, $\bar{f}_{\textit{text}}$, $\bar{f}_{\textit{key}}$, and $\bar{f}_{\textit{red}}$. Among them, $\bar{f}_{\textit{total}}$ is the overall normalized frequency obtained by counting all tokens (i.e., applying Eq.~\eqref{eq:freq-norm} to the full token set without partitioning), while $\bar{f}_{\textit{text}}$, $\bar{f}_{\textit{key}}$, and $\bar{f}_{\textit{red}}$ are the normalized frequencies computed on the corresponding token subsets.
    \item In Sec.~\ref{sec:method}, since our analysis has already focused on text tokens and key vision tokens, we adopt the more compact notation $\bar{f}_t$ and $\bar{f}_v$ to denote the normalized frequencies of the two modalities. Specifically, $\bar{f}_t$ denotes the frequency computed on text tokens (identical to $\bar{f}_{\textit{text}}$ in Sec.~\ref{sec:concept_verification}), and $\bar{f}_v$ denotes the frequency computed on key vision tokens only (identical to $\bar{f}_{\textit{key}}$); the redundant vision tokens are no longer involved.
\end{itemize}

\subsection{Cross-Dataset Cosine Similarity}
\label{app:freq-cosine}

To measure how similar the routing behavior is between two calibration sets $\mathcal{D}_a$ and $\mathcal{D}_b$, we use the normalized frequency vectors above. For a given token type $m$, we first compute the cosine similarity at every layer:
\begin{equation}
s^{(\ell)}_m \;=\; \cos\!\big(\bar{\boldsymbol{f}}^{(\ell)}_{m}(\mathcal{D}_a),\; \bar{\boldsymbol{f}}^{(\ell)}_{m}(\mathcal{D}_b)\big),
\label{eq:cos-layer}
\end{equation}
and then average across all $L$ MoE layers to obtain the final similarity:
\begin{equation}
S_m(\mathcal{D}_a, \mathcal{D}_b) \;=\; \frac{1}{L} \sum_{\ell=1}^{L} s^{(\ell)}_m.
\label{eq:cos-final}
\end{equation}
Since each $\bar{\boldsymbol{f}}^{(\ell)}_{m}$ is a probability distribution with non-negative entries, every per-layer similarity satisfies $s^{(\ell)}_m \in [0, 1]$, and consequently $S_m(\mathcal{D}_a, \mathcal{D}_b) \in [0, 1]$. A value closer to $1$ indicates that, for token type $m$, the two datasets induce more consistent expert selection patterns throughout the network, supporting the cross-dataset transferability claim in the main text.


\section{Performance on Language-Only Tasks}
\label{app:language-only}

\begin{table*}[t]
\centering
\caption{Performance on language-only benchmarks. Our method preserves language capabilities with limited degradation at both bit-widths. Qwen3-VL-30B-A3B and Kimi-VL-A3B are short for Qwen3-VL-30B-A3B-Instruct and Kimi-VL-A3B-Instruct, respectively. ARC-E and ARC-C denote ARC-Easy and ARC-Challenge.}
\label{tab:language-only}
\resizebox{\textwidth}{!}{
\begin{tabular}{c|ccccccc|c}
\toprule
\textbf{Qwen3-VL-30B-A3B} & ARC-E & ARC-C & HellaSwag & LAMBADA & MMLU & PIQA & WinoGrande & Avg \\
\midrule
Baseline (FP16) & 80.56 & 54.69 & 62.03 & 71.49 & 80.75 & 80.25 & 74.59 & 70.63 \\
W3A16 & 79.92 & 52.47 & 58.67 & 68.45 & 75.89 & 78.84 & 71.01 & 67.89 \\
W2A16 & 77.48 & 49.23 & 53.12 & 62.32 & 65.38 & 75.57 & 66.48 & 64.23 \\
\midrule
\textbf{Kimi-VL-A3B} & ARC-E & ARC-C & HellaSwag & LAMBADA & MMLU & PIQA & WinoGrande & Avg \\
\midrule
Baseline (FP16) & 82.62 & 53.92 & 59.57 & 71.61 & 69.01 & 80.20 & 71.59 & 69.79 \\
W3A16 & 80.72 & 52.13 & 56.56 & 70.95 & 64.71 & 78.89 & 70.32 & 67.75 \\
W2A16 & 75.67 & 42.92 & 48.14 & 63.09 & 57.46 & 76.39 & 64.64 & 61.19 \\
\bottomrule
\end{tabular}
}
\end{table*}

The core of our method is to jointly consider the cross-modal expert importance bias between language and vision modalities and the intra-vision expert importance bias between key and redundant vision tokens. In the main experiments, we primarily focus on evaluating quantized MoE-MLLMs on multimodal benchmarks, where our method significantly outperforms other PTQ methods designed for MLLMs, MoE-LLMs, or MoE-MLLMs. Meanwhile, it is important to ensure that our quantization does not severely degrade the language-only capabilities of MoE-MLLMs.

To this end, we evaluate our method on Qwen3-VL-30B-A3B-Instruct and Kimi-VL-A3B-Instruct at both 2-bit and 3-bit average bit-widths across seven widely used language-only benchmarks: ARC-Easy, ARC-Challenge, HellaSwag, LAMBADA, MMLU, PIQA, and WinoGrande. Results are reported in Table~\ref{tab:language-only}.

As shown in Table~\ref{tab:language-only}, at 3-bit average bit-width, the average performance degradation is limited to approximately 3\% on Qwen3-VL-30B-A3B-Instruct and 2\% on Kimi-VL-A3B-Instruct. These results confirm that our modality-aware mixed-precision quantization method effectively preserves the language-only capabilities of MoE-MLLMs while substantially improving multimodal performance as demonstrated in the main experiments.

\section{Quantizing Both Vision Module and LLM}
\label{app:quant-vision}

In the main experiments, we focus on quantizing the LLM part of MoE-MLLMs while keeping the vision module (i.e., the vision encoder and the multi-modal projector, plus the merger module for Qwen3-VL-30B-A3B-Instruct) at original precision. We adopt this setting because the vision module accounts for only a negligible fraction of the total weight memory. As reported in Table~\ref{tab:vision-llm-memory}, the vision module occupies merely around 1\,GB of memory on both Qwen3-VL-30B-A3B-Instruct and Kimi-VL-A3B-Instruct, which is less than 3\% of the total model size. Therefore, keeping the vision module at original precision does not noticeably affect the overall memory footprint. Furthermore, our method specifically targets the mixed-precision quantization of MoE experts in MoE-MLLMs and is orthogonal to methods that quantize the vision module, such as Speed-Q.

\begin{table*}[htb]
\centering
\caption{Weight memory footprint of the vision module versus the LLM part on Qwen3-VL-30B-A3B-Instruct and Kimi-VL-A3B-Instruct. The vision module accounts for less than 3\% of the total weight memory on both models.}
\label{tab:vision-llm-memory}
\begin{tabular}{c|cc|cc|c}
\toprule
\textbf{Model} & \textbf{Vision (GB)} & \textbf{Vision (\%)} & \textbf{LLM (GB)} & \textbf{LLM (\%)} & \textbf{Total (GB)} \\
\midrule
Qwen3-VL-30B-A3B-Instruct & 1.077 & 1.73\% & 61.064 & 98.27\% & 62.142 \\
Kimi-VL-A3B-Instruct      & 0.895 & 2.73\% & 31.920 & 97.27\% & 32.815 \\
\bottomrule
\end{tabular}
\end{table*}

Nevertheless, to provide more comprehensive results, we further evaluate the setting where both the vision module and the LLM part are quantized. For the LLM part, we apply our mixed-precision expert quantization method as in the main experiments, with average bit-widths of 3-bit and 2-bit. For the vision module, we simply apply GPTQ to quantize it to 4-bit. Evaluation is conducted on the same 10 multimodal benchmarks as in the main experiments, and results are summarized in Table~\ref{tab:joint-vision-llm}. In the table, the column ``Vision Q.'' indicates whether the vision module is additionally quantized to 4-bit ($\checkmark$) or kept at original precision ($\times$).

\begin{table*}[t]
\centering
\caption{Performance comparison between LLM-only quantization and joint quantization of both the vision module and the LLM. The ``Vision Q.'' column indicates whether the vision module is quantized to 4-bit using GPTQ ($\checkmark$) or kept at original precision ($\times$). Qwen3-VL-30B-A3B and Kimi-VL-A3B are short for Qwen3-VL-30B-A3B-Instruct and Kimi-VL-A3B-Instruct, respectively.}
\label{tab:joint-vision-llm}
\resizebox{\textwidth}{!}{
\begin{tabular}{c|c|cccccccccc|c}
\toprule
\textbf{Qwen3-VL-30B-A3B} & \textbf{Vision Q.} & ChartQA & MMBench & MME-R & MMMU & MMStar & POPE & GQA & TextVQA & InfoVQA & VizWiz & Avg \\
\midrule
Baseline (BF16) & $\times$      & 85.20 & 86.86 & 52.48 & 52.56 & 60.05 & 89.92 & 62.96 & 83.37 & 81.46 & 71.64 & 72.65 \\
MODE-W3          & $\times$      & 83.00 & 82.56 & 51.77 & 50.33 & 55.68 & 89.44 & 60.21 & 80.65 & 77.62 & 68.82 & 70.01 \\
MODE-W3          & $\checkmark$  & 81.92 & 80.59 & 48.40 & 46.56 & 52.41 & 89.04 & 59.47 & 80.09 & 77.13 & 68.00 & 68.36 \\
MODE-W2          & $\times$      & 73.11 & 75.71 & 44.85 & 44.63 & 45.34 & 87.28 & 56.99 & 75.45 & 66.19 & 63.39 & 63.29 \\
MODE-W2          & $\checkmark$  & 74.59 & 72.96 & 41.72 & 41.52 & 42.26 & 86.51 & 56.33 & 75.98 & 66.47 & 59.34 & 61.77 \\
\midrule
\textbf{Kimi-VL-A3B} & \textbf{Vision Q.} & ChartQA & MMBench & MME-R & MMMU & MMStar & POPE & GQA & TextVQA & InfoVQA & VizWiz & Avg \\
\midrule
Baseline (BF16) & $\times$      & 89.44 & 82.99 & 44.45 & 52.33 & 49.41 & 87.17 & 62.72 & 88.69 & 83.61 & 70.59 & 71.14 \\
MODE-W3          & $\times$      & 87.26 & 80.84 & 41.59 & 48.51 & 48.66 & 87.10 & 61.11 & 86.24 & 79.87 & 69.42 & 69.06 \\
MODE-W3          & $\checkmark$  & 86.78 & 80.50 & 42.16 & 48.31 & 46.51 & 87.37 & 60.39 & 85.65 & 78.80 & 69.02 & 68.55 \\
MODE-W2          & $\times$      & 75.20 & 75.23 & 38.71 & 45.11 & 44.11 & 85.46 & 58.30 & 79.38 & 67.75 & 64.52 & 63.38 \\
MODE-W2          & $\checkmark$  & 67.04 & 68.45 & 38.81 & 43.77 & 38.79 & 85.53 & 58.45 & 77.39 & 65.46 & 65.09 & 60.88 \\
\bottomrule
\end{tabular}
}
\end{table*}

As shown in Table~\ref{tab:joint-vision-llm}, on Qwen3-VL-30B-A3B-Instruct, quantizing the vision module to 4-bit yields an additional memory saving of approximately $1.077 \times 3/4 \approx 0.81$\,GB, while incurring about 1.7\% and 1.5\% relative average accuracy drops under the 3-bit and 2-bit LLM settings, respectively. On Kimi-VL-A3B-Instruct, the joint quantization is fairly robust under the 3-bit setting with only a 0.5\% relative drop, but the degradation becomes more pronounced under the 2-bit setting, exceeding 2.5\%. Although our method is not specifically designed for the vision module, we hope these results can serve as a useful reference for future works on jointly quantizing both modalities of MoE-MLLMs.

\section{The Importance of Protecting Attention Module at 4-bit}
\label{app:attn-protection}

In our experiments, we observe that for MoE-MLLMs, although the attention module accounts for only a small fraction of the total parameters, it is critical to the overall performance, especially under extremely low-bit quantization. For instance, under 2-bit quantization, quantizing the MoE layers to 2-bit causes noticeable accuracy degradation but does not lead to model collapse, whereas directly quantizing the attention module to 2-bit causes the model to collapse entirely. In contrast, keeping the attention module at 4-bit substantially mitigates the quantization loss. We attribute this to the fact that in MoE-MLLMs, experts in the MoE layers are sparsely activated, while the attention module operates on every token, making it far more sensitive to quantization noise. While we do not regard this as a novel contribution, we find it highly valuable for practical implementation and deployment, and therefore discuss it in detail here.

\paragraph{Parameter footprint of the attention module.}
We first show that the attention module occupies only a small fraction of the LLM parameters. As reported in Table~\ref{tab:attn-moe-memory}, the attention module accounts for less than 3\% of the LLM weight memory on both Qwen3-VL-30B-A3B-Instruct and Kimi-VL-A3B-Instruct, while the MoE layers dominate the parameter budget at over 93\%.

\begin{table*}[htb]
\centering
\caption{Weight memory footprint of the attention module versus the MoE layers in the LLM part of Qwen3-VL-30B-A3B-Instruct and Kimi-VL-A3B-Instruct. The attention module accounts for less than 3\% of the LLM weight memory on both models, while the MoE layers dominate at over 93\%.}
\label{tab:attn-moe-memory}
\resizebox{\textwidth}{!}{
\begin{tabular}{c|cc|cc|c}
\toprule
\textbf{Model} & \textbf{Attention (GB)} & \textbf{Attention (\%)} & \textbf{MoE Layers (GB)} & \textbf{MoE Layers (\%)} & \textbf{LLM Total (GB)} \\
\midrule
Qwen3-VL-30B-A3B-Instruct & 1.812 & 2.97\% & 58.007 & 94.99\% & 61.064 \\
Kimi-VL-A3B-Instruct      & 0.743 & 2.33\% & 29.696 & 93.03\% & 31.920 \\
\bottomrule
\end{tabular}
}
\end{table*}

\begin{table*}[t]
\centering
\caption{Comparison between fully quantizing the LLM to 2-bit (GPTQ-W2) and protecting the attention module at 4-bit while quantizing the MoE layers to 2-bit (GPTQ-W2-Attn4). Protecting the attention module at 4-bit substantially mitigates the quantization loss on both models.}
\label{tab:attn-protection}
\resizebox{\textwidth}{!}{
\begin{tabular}{c|cccccccccc|c}
\toprule
\textbf{Qwen3-VL-30B-A3B-Instruct} & ChartQA & MMBench & MME-R & MMMU & MMStar & POPE & GQA & TextVQA & InfoVQA & VizWiz & Avg \\
\midrule
Baseline (BF16) & 85.20 & 86.86 & 52.48 & 52.56 & 60.05 & 89.92 & 62.96 & 83.37 & 81.46 & 71.64 & 72.65 \\
GPTQ-W2         & 31.12 & 22.46 & 12.94 & 25.56 & 30.48 & 50.84 & 36.89 & 40.98 & 39.07 & 37.55 & 34.99 \\
GPTQ-W2-Attn4      & 58.16 & 52.68 & 32.91 & 36.11 & 36.08 & 79.56 & 45.44 & 56.38 & 59.07 & 52.36 & 50.88 \\
\midrule
\textbf{Kimi-VL-A3B-Instruct} & ChartQA & MMBench & MME-R & MMMU & MMStar & POPE & GQA & TextVQA & InfoVQA & VizWiz & Avg \\
\midrule
Baseline (BF16) & 89.44 & 82.99 & 44.45 & 52.33 & 49.41 & 87.17 & 62.72 & 88.69 & 83.61 & 70.59 & 71.14 \\
GPTQ-W2         & 27.62 & 23.44 & 24.23 & 27.33 & 23.01 & 70.29 & 37.09 & 48.02 & 22.42 & 25.20 & 32.87 \\
GPTQ-W2-Attn4      & 45.64 & 50.84 & 29.71 & 34.00 & 32.66 & 73.10 & 46.11 & 64.24 & 49.27 & 43.42 & 46.90 \\
\bottomrule
\end{tabular}
}
\end{table*}

\paragraph{Effect of protecting attention at 4-bit.}
To empirically verify the importance of protecting the attention module, we use GPTQ as a representative quantization method and compare two settings on Qwen3-VL-30B-A3B-Instruct and Kimi-VL-A3B-Instruct: (i) \textbf{GPTQ-W2}, where all components of the LLM (including both the attention module and the MoE layers) are quantized to 2-bit; and (ii) \textbf{GPTQ-W2-Attn4}, where only the MoE layers are quantized to 2-bit while the attention module is kept at 4-bit. Both settings are evaluated on the same 10 multimodal benchmarks as in the main experiments, and the results are summarized in Table~\ref{tab:attn-protection}. GPTQ-W2 denotes uniform 2-bit quantization applied to both attention and MoE layers, whereas GPTQ-W2-Attn4 keeps attention at 4-bit precision while quantizing the MoE layers to 2-bit.

As shown in Table~\ref{tab:attn-protection}, under the pure 2-bit setting, Qwen3-VL-30B-A3B-Instruct exhibits clear performance collapse on multiple benchmarks including ChartQA, MME-R, TextVQA, and InfoVQA, with accuracy on MME-R dropping to as low as 12.94. Once the attention module is protected at 4-bit, the performance is largely recovered, with the average accuracy improving from 34.99 to 50.88 (a relative gain of over 45\%). A similar trend is observed on Kimi-VL-A3B-Instruct, where the average accuracy increases from 32.87 to 46.90. These results empirically confirm that protecting the attention module at 4-bit is critical under extremely low-bit quantization.

Moreover, the additional memory cost of protecting the attention module is marginal. Taking Qwen3-VL-30B-A3B-Instruct as an example, the extra memory introduced by promoting attention from 2-bit to 4-bit can be compensated by reducing the bit-width of merely 8 experts (out of $128 \times 48$ experts in total) by 1 bit across all layers. Given that the experts are empirically far more robust to quantization than attention, protecting the attention module at 4-bit is a highly cost-effective design choice and is strongly recommended in practice.

\section{Multi-GPU Parallel GPTQ for MoE-MLLMs}
\label{app:parallel-gptq}

\subsection{Motivation and Design}
Fine-grained MoE-MLLMs typically contain hundreds of routed experts per layer, each requiring independent GPTQ~\citep{frantar2023gptqaccurateposttrainingquantization} quantization of its \texttt{gate\_up\_proj} and \texttt{down\_proj} matrices. Conventional GPTQ pipelines traverse these experts strictly one after another within each MoE layer, and as the number of experts grows, the cumulative cost of this per-expert sequential traversal makes MoE-layer calibration the dominant bottleneck of the entire quantization procedure on fine-grained MoE models. Since the Hessian construction and weight update for different experts are, however, mutually independent, the expert-quantization stage is in fact an \emph{embarrassingly parallel} workload. We therefore introduce an expert-level multi-GPU scheme that follows a \textit{``serial collection $\to$ parallel quantization $\to$ serial propagation''} pattern: inter-layer ordering, attention quantization, calibration forward passes, activation/Hessian collection, and the post-quantization re-forward used for layer-wise error propagation all remain strictly serial; only the inner loop over experts within a single quantization stage is dispatched concurrently across GPUs. This scheduling change leaves the GPTQ mathematics, the quantization order, and all inter-layer dependencies untouched, so the output is bit-identical to the serial baseline. Algorithmic equivalence is guaranteed by four invariants: (i)~layer order is preserved, so layer $\ell\!+\!1$ always sees inputs produced by the fully quantized layer $\ell$; (ii)~within a layer, \texttt{down\_proj} quantization strictly follows the re-forward of the quantized \texttt{gate\_up\_proj}, satisfying GPTQ's sequential-quantization requirement; (iii)~experts within the same stage are mathematically independent; and (iv)~attention submodules and routers are quantized exactly as in the serial baseline.

\subsection{Implementation}
We take a single projection matrix of a single expert as the minimal task unit, yielding up to $2E$ independent tasks per MoE layer with $E$ experts. Tasks are assigned to $G$ GPUs in a round-robin manner, and each GPU processes its local queue sequentially while the $G$ queues run concurrently; multi-threading is sufficient to drive this parallelism, since CUDA kernels release the Python GIL during execution. Each worker independently transfers the assigned expert weight and its corresponding inputs to its target GPU, performs the GPTQ update locally, and offloads the quantized result back to CPU before releasing its GPU memory. The main thread, residing on a single primary device, retains responsibility for calibration, activation collection, and re-forward propagation between stages, so that quantized expert weights produced in parallel are gathered and reinstalled into the model before the next dependent stage begins. To bound peak memory, expert inputs and Hessians are cloned into per-task structures before dispatch and released immediately afterwards, and explicit cache cleanup and synchronization are performed at the end of each layer. The scheme is implemented as a drop-in replacement for the standard GPTQ entry point and is agnostic to the specific MoE weight organization adopted by different model families.

\subsection{Complexity and Speedup}
Let $E$ be the number of routed experts per layer, $G$ the number of GPUs, and $t$ the GPTQ time of a single expert projection. The serial baseline costs $\mathcal{O}(E\cdot t)$ per stage, while our scheme reduces this to $\mathcal{O}(\lceil E/G\rceil\cdot t)$, giving an expert-stage speedup of $\min(E, G)$. Denoting the wall-clock time of all serial components by $T_s$ and the serial cost of expert quantization by $T_p$, the overall speedup can be expected as:
\[
\mathrm{Speedup} \;=\; \frac{T_s + T_p}{T_s + T_p / G},
\]
which approaches $G$ when $T_p \gg T_s$, as is typical for fine-grained MoE-MLLMs with large $E$. In our experiments, Kimi-VL-A3B-Instruct contains $64$ routed experts per layer, while Qwen3-VL-30B-A3B-Instruct and InternVL3.5-30B-A3B each contain $128$ experts per layer, placing all three models well within the regime where the parallel speedup approaches $G$.

\section{LLM Usage}
In the preparation of this manuscript, large language models were used to polish the writing, including grammar refinement, phrasing improvement, and minor stylistic edits.

\end{document}